\documentclass{article}

\usepackage{changes}
\usepackage{amsmath,amssymb}
\usepackage{subcaption}

\usepackage{algorithm2e}
\usepackage{algpseudocode}
\RestyleAlgo{ruled}
\usepackage{hyperref}

\usepackage{todonotes}

\RequirePackage[left=2.7cm,
right=2.7cm,
top=1.8cm,
bottom=1.5cm]{geometry}

\title{
		Designing Robust Biotechnological Processes Regarding Variabilities using Multi-Objective Optimization Applied to a Biopharmaceutical Seed Train Design
}
\author{Tanja Hern\'{a}ndez Rodr\'{i}guez,
Anton Sekulic,
Markus Lange-Hegermann,\\
Bj\"orn Frahm
}

\date{}

\begin{document}
	
\maketitle

\abstract{Development and optimization of biopharmaceutical production processes with cell cultures is cost- and time-consuming and often performed rather empirically. Efficient optimization of multiple-objectives like process time, viable cell density, number of operating steps \& cultivation scales, required medium, amount of product as well as product quality depicts a promising approach. This contribution presents a workflow which couples uncertainty-based upstream simulation and Bayes optimization using Gaussian processes. Its application is demonstrated in a simulation case study for a relevant industrial task in process development, the design of a robust cell culture expansion process (seed train), meaning that despite uncertainties and variabilities concerning cell growth, low variations of viable cell density during the seed train are obtained. Compared to a non-optimized reference seed train, the optimized process showed much lower deviation rates regarding viable cell densities (\textless~10\% instead of 41.7\%) using 5 or 4 shake flask scales and seed train duration could be reduced by 56\,h from 576\,h to 520\,h. Overall, it is shown that applying Bayes optimization allows for optimization of a multi-objective optimization function with several optimizable input variables and under a considerable amount of constraints with a low computational effort. This approach provides the potential to be used in form of a decision tool, e.g.\ for the choice of an optimal and robust seed train design or for further optimization tasks within process development.}

\section{Introduction}

The development and optimization of biopharmaceutical production processes with cell cultures is cost- and time-consuming, requiring substantial lab work. This necessitates thorough planning of experiments and processes, taking into account existing process knowledge.
The need of model-based decision support in biopharmaceutical manufacturing has been emphasized by the US Food and Drug Administration (FDA) \cite{Herwig.2015,U.S.DepartmentofHealthandHumanServicesFoodandDrugAdministration.}, including taking into account available prior know-how and experience within the decision process and uncertainties \cite{Sokolov.2020}.
Such methods are still not state of the art for cell culture processes during development or manufacturing \cite{Sokolov.2020, Xie.2019}, although first approaches have been proposed, for example for titer optimization of a mammalian cell culture process \cite{Liu.2017}.
This highlights a need for improved methods and tools for optimal experimental design, optimal and robust process design and process optimization for the purposes of monitoring and controlling during manufacturing.

Optimization for \emph{one} objective criterion (e.g.\ final titer) is relatively straight forward, i.e.\ building an objective function with a unique response variable and applying an appropriate optimization algorithm to maximize this function.
However, in industry, it is typically desired to optimize \emph{several} conflicting objectives at a time, leading to suitable trade-offs and compromises. For example when trying to maximize final titer via viable cell density while minimizing cultivation time.
Multi-objective optimization provides a decision-making tool for optimal decisions in the presence of trade-offs between two or more conflicting criteria.

However, multi-objective optimization is more challenging.
Its application is still not state of the art in the context of cell culture processes, probably due to a lack of related studies and instructions. Moreover, within the manufacturing life cycle of biopharmaceuticals, some phases are better investigated than others. Still very few investigations are reported concerning the cell expansion process (called seed train). It consists of several consecutive cultivation and passaging (transfer) steps, starting with a small amount of cell suspension because cells are frozen in small vials until they are used for a production process. The goal is to expand the number of viable cells in order to reach the required amount to inoculate (start) the production bioreactor (e.g.\ 10,000~L at industrial scale) while keeping them in a healthy and growing state. A high amount of operational requirements and constraints have to be fulfilled and, as reported in literature \cite{Le.2012,Bohl.2020}, the cell expansion process critically effects product quality and the amount of product at production scale. In \cite{Bohl.2020} for example, the passage duration as well as the initial viable cell density for each passage are reported as important parameters with high impact on process time and productivity at production scale. A careful and optimal planning of a seed train is therefore essential. However, this is not a trivial task due to the inherent variability concerning cell growth (cell growth differs from cell line to cell line and also from cultivation run to cultivation run) and uncertainty about the real state of the process due to considerable measurement uncertainties. This requires the design a reproducible process which is robust regarding viable cell density, meaning that despite (initial) variabilities concerning cell growth, low variations of viable cell density at the end of the seed train are obtained.
The goal of this paper is to close the gap between state of the art optimization techniques and modern techniques from machine learning to improve the biopharmaceutical production by allowing easy to use yet powerful multi-objective optimization.

%

In most multi-objective optimization problems no single best (unique optimal) solution exist, instead there is a set of optimal solutions (also called Pareto optimal solutions or non-dominated solutions), meaning for each solution that one criterion cannot be improved without degrading at least one of the other criteria. So, the decision maker has to choose from the set of non-dominated solutions according to the most preferred or important objective criterion. 
A promising approach to optimize objective functions, which are expensive to evaluate, is Bayes optimization.
The methodology of Bayes optimization dates back to the work of Harold Kushner in 1964 \cite{Kushner.1964} and gained impact through the work of Jones et al.\ in 1998 \cite{Jones.1998}.
It is a probabilistic global optimization method for finding the maximum of expensive to evaluate objective functions or unknown (black-box) objective functions that are approximated using simulations \cite{Brochu.12.12.2010}.

In practice, the objective function could be the outcome of interest of a process, for example process productivity or control metrics to describe the quality of a product.
Input parameters can be process parameters needed to be optimized.
Bayesian optimization \cite{Shahriari.2016} creates a quick to evaluate model, the so-called \emph{surrogate model}, of the objective function. In order to reduce the objective function evaluations, the surrogate model is iteratively trained and updated on new data.
The positions of this new data are chosen by finding trade-off between exploration (improving the surrogate model) and exploitation (finding optimal points).
Typical surrogate models are Gaussian processes.

Gaussian processes (GP's) are popular machine learning models \cite{Rasmussen.2008}, because due to their Bayesian nature, they work well with few data points \cite{tulsyan2019industrial}.
Furthermore, they allow to include expert knowledge \cite{LH_AlgorithmicLinearlyConstrainedGaussianProcesses,LH_AlgorithmicLinearlyConstrainedGaussianProcessesBoundaryConditions} and can be used in dynamic systems \cite{Bradford.2020, Bradford.2018b}.
GP's are very flexible non-parametric models, hence they can approximate any function and do not assume a predefined set of modeling functions.

Bayes optimization is successfully applied in many fields of research and economics \cite{Petsagkourakis.04.06.2020}. 
Also applications of Bayes optimization in the field of bioprocess engineering were published during the last decade \cite{Bradford.2018,Bradford.2018b,Clayton.2020,Narayanan.2022}.
Furthermore, this methodology was shown to be efficient in solving multi-objective optimization problems \cite{Yang.2019} and has also be applied for parameter estimation of kinetic parameters \cite{Manheim.2019}. 
However, no applications are reported so far, applying model-based multi-objective Bayes optimization within biopharmaceutical process development.

This contribution aims to present the concept of a workflow which couples uncertainty-based upstream simulation and Bayes optimization using Gaussian processes and its application in form of a simulation case study to illustrate its applicability to a relevant industrial task in process development.\\
This simulation case study addresses the question if a reference seed train setup comprising 5 shake flask scales can be optimized through varying shake flask volumes and
how many shake flask scales, three, four or five, are recommendable in terms of two objective criteria, seed train duration and deviation rate. 
Moreover it is investigated how the results change if cells grow with 5\% lower or 5\% higher maximum cell-specific growth rate. \\
Afterwards, two more objective criteria, titer (product concentration) and viability after 8 days in the production bioreactor, are added and seed train optimization is performed regarding four objective criteria simultaneously.\\
Furthermore, the suitability of the proposed method and the required number of iterations is evaluated with respect to the obtained information gain. 

%

\section{Methods}
The main components of the applied methodology and the corresponding tools are described.

\subsection{Upstream simulation}\label{method: seed train simulation}
Upstream simulation comprises simulation of the cell expansion process (seed train) and simulation of the production scale.
The reference upstream process taken as an application example for the here presented simulation case study comprises five consecutive shake flask scales followed by three bioreactor scales and one production scale, similar to the upstream process investigated in \cite{HernandezRodriguez.2019}. Further specifications are listed in Table~\ref{tab: setup_bayes_opt_ex_2}. \\
A mathematical model is required, describing cell growth and interactions with the main limiting substrates and eventually inhibiting metabolites over time.
A cell growth model, a system of ordinary differential equations (ode), already adapted to an industrial cell culture upstream process using a CHO cell line \cite{HernandezRodriguez.2019} has been used, which describes the dynamic behavior of viable and total cell density, $X_{\rm v}$ and $X_{\rm t}$, concentrations of glucose $c_{\rm Glc}$, glutamine $c_{\rm Gln}$, lactate $c_{\rm Lac}$, ammonia $c_{\rm Amm}$ and product (volumetric titer) $c_{\rm titer}$ (see Table~\ref{tab: model_eq_novartis} in the supplementary material). 

Moreover, such an upstream process includes several constraints, operation steps and process parameters (e.g.\ concerning passaging intervals, substrate/nutrient concentrations, initial viable cell densities and viable cell densities before transferring cells into the next cultivation vessel as well as the amount of cell suspension and fresh medium), which have to be considered in the simulation workflow.
A detailed description of the required components and calculation routines are described in \cite{HernandezRodriguez.2020b,HernandezRodriguez.2020}. 

Besides these requirements, several passaging strategies can be applied, helping to decide at which point in time cells should be transferred from one cultivation vessel into the next larger one and how to perform these passaging steps (e.g.\ which amount of cell suspension should be mixed with how much fresh cell culture medium). 

For the here presented simulation study, the passaging strategy for robust seed train design was chosen, where robustness refers to the reproducibility of the seed train regarding viable cell density, meaning that despite initial uncertainties and variabilities concerning cell growth, low variations of viable cell density at the end of the seed train are obtained. This strategy grounds on the objective of reaching the previously determined threshold of viable cell density and corresponding probability distributions of viable cell density at different points in time. These distributions are used in combination with a utility function following the mean-variance principle: The utility function $U(t)$ is defined as a function of viable cell density $X_{\rm v}$ including the expected value ${\rm E}(X_{\rm v})$ and the variance  ${\rm Var}(X_{\rm v})$ of viable cell density as well as a risk aversion parameter $\alpha$ which controls the amount of risk (amount of uncertainty) the user is willing to bear. In the here presented example, the risk refers to the probability that viable cell density differs from the expected value (predicted mean). A risk aversion value of $\alpha\,=\,1$ would mean that the expected time profile minus one time the standard deviation of $X_{\rm v}$ is considered.

The utility function is defined through:
\begin{equation}\label{eq: utility}
	U(t) = E(X_{\rm v}(t)) - \alpha \sqrt{{\rm Var}({X_{\rm v}(t)})}
\end{equation}
Based on the simulated time profiles of the current cultivation scale (by solving the corresponding ode system), Eq.~\ref{eq: utility} is used to calculate the utility function value $U(t)$ per hour and to check if this value reaches or exceeds the required transfer viable cell density $X_{\rm v,transfer}$ which is necessary to inoculate (start) the next cultivation scale fulfilling the required seeding (initial) viable cell density and the filling volume. 

In a next step, it is evaluated if the calculated point in time lies within the range of practically feasible points in time for cell passaging, $T_{\rm p}$.  
Thus, the objective is to find the minimum point in time out of the set of practically feasible points in time for passaging, $T_{\rm p}$, which fulfills

\begin{align}
	&U(t) \ge  X_{\rm v,transfer}, \\
	&\mbox{subject to: \quad} t\in T_{\rm p}.
\end{align}

Based on the obtained point in time and the corresponding concentrations of viable cells, total cells, substrates and metabolites at this point in time, starting concentrations (=\,initial values of the system of ordinary differential equations) of the next cultivation scale are calculated based on the defined configurations and constraints (e.g.\ working volumes, acceptable range of seeding viable cell density and medium concentrations). 
This calculation has to be performed for every cultivation scale and passaging step. For more details refer to \cite{HernandezRodriguez.2019, HernandezRodriguez.2020, HernandezRodriguez.2020b}.

{\small 
	\begin{table}[h]
		\centering
		\renewcommand{\arraystretch}{1.0} 
		\caption[Specification of the exemplary seed train setup.]{Specification of the exemplary seed train setup providing information concerning cultivation vessels, required viable cell densities and the transfer of cells from one cultivation vessel into the next larger one, assumed in this work.}
		\label{tab: setup_bayes_opt_ex_2}
		\begin{tabular}{l|l}
			\hline
			Seed train setup & \\
			\hline\\[-1.0em] 
			Flask scales: & 3, 4 or 5 flask scales between 0.014\,L and 8\,L filling volume \\[1.5mm]
			Bioreactor scales: & 3 bioreactor scales, 38 L, 302 L and  2054 L filling volume \\[1.5mm]
			Production bioreactor: & 9500 L filling volume \\[1.5mm]
			Optimal range for & \\
			viable seeding cell density: & $3\cdot\,10^{8}$ -- $3.5\cdot\,10^{8}$ cells L$^{-1}$ ($3\cdot\,10^{5}$ -- $3.5\cdot\,10^{5}$ cells mL$^{-1}$) \\ [1.5mm]
			Optimal range for & \\
			transfer viable cell density: & $0.1\cdot\,10^{10}$ -- $1\cdot\,10^{10}$ cells L$^{-1}$  ($0.1\cdot\,10^{7}$ -- $1\cdot\,10^{7}$ cells mL$^{-1}$ ) \\[1.5mm]
			Target seeding (initial)  & \\
			viable cell density:      & 3.15 $\cdot\,10^{8}$ cells L$^{-1}$ (3.15 $\cdot\,10^{5}$ cells~mL$^{-1}$)  \\
						     & (= minimum viable seeding VCD + 5\%) \\
			\hline \\[-1.0em]
			Strategy concerning point &  'Xv transfer', i.e.\ passaging as soon as the calculated \\
			in time for cell passaging: & required viable transfer cell density is reached  \\[1.5mm]
			Practically feasible & Passaging between 48 and 120 h possible \\
			points in time for passaging: & (flexible ranges)  \\[1.5mm]
			Strategy concerning & Discard cell suspension during the passaging step, if\\
			current and new volume: & required to start within an optimal seeding \\
			&  cell density range\\
			\hline
		\end{tabular}
	\end{table}
}


\subsection{Bayes optimization}\label{method: Bayes_optimization}

A typical mathematical optimization problem is the following:
Given an objective function $f: \mathcal{X} \rightarrow \mathbb{R}$ over input space $\mathcal{X}\subseteq\mathbb{R}^d$, the aim is to find an argument $x^*\in\mathcal{X}$, which optimizes (minimizes or maximizes) $f$.

The idea behind Bayes Optimization consists in creating a simple, probabilistic and cheap to evaluate model, a so-called \emph{surrogate model} (substitute model), of the objective function $f$ \cite{Rasmussen.2008,Frazier.08.07.2018,Brochu.12.12.2010}. 
Bayesian optimization reduces the number of evaluations of the objective $f$ via the following iterative approach: 
Before sampling $f$ at another point, we take into account a trade-off between exploration (i.e.\ sampling of areas of high uncertainties) and exploitation (sampling from areas which are likely to move towards the optimum), which is encoded in a so-called acquisition function.
We can find such points quickly from evaluation of the surrogate model.

Within Bayes optimization the following steps are performed:
\begin{enumerate}
	\item Generate a set of initial points and evaluate the objective function at these points.
	\item Train the surrogate model based on all evaluated points. 
	\item Optimize the acquisition function, which determines the next candidate point $x_c$ to be evaluated.
	\item Compute $f(x_c)$, the objective function $f$ at the candidate point $x_c$.
	\item Repeat steps 2-4 for $N$ iterations
\end{enumerate}

The key of Bayesian optimization is not to rely on local approximations as many other optimization algorithms and instead to have a global viewpoint of also evaluating the function at unknown positions. 

The acquisition function is used to propose the next candidate point to be evaluated based on specific criteria, for example the expected improvement of the optimization criteria, and on the reduction of predictive uncertainty. 
As in the case of the kernels, there is also a wide variety of possible acquisition functions to choose from. In this study, the \emph{Expected Improvement} (EI) acquisition function is used \cite{Couckuyt.2014, AntonSekulic.2020}.

Gaussian processes (GPs) are well suited surrogate models when making few assumptions \cite{Brochu.12.12.2010}.
Just like a Gaussian distribution (a normal probability distribution) is fully described by its mean $m$ and variance $\sigma^2$, a GP is fully
described by a mean function $m(x)$ and a covariance function $k(x;x^{'})$ \cite{Rasmussen.2008}.
A GP is an extension of a multivariate Gaussian (or normal) distribution to distributions of functions in the sense that if a function $y$ follows a GP distribution, i.e.\ $y\sim \mathcal{GP}\left( m, k \right)$, then every evaluation of the function follows a Gaussian distribution $y(x) \sim \mathcal{N}\left( m(x), k(x,x) \right)$.
In particular, a GP returns mean and variance of the possible function values (instead of just returning a scalar), and hence also provides information about the uncertainty of a prediction.
Moreover, GPs can take into account uncertainty in form of noise, the class of Gaussian processes is closed under Bayesian updates, and such updates are computationally tractable \cite{Snoek.13.06.2012}.

The covariance function describes the assumed characteristics such as smoothness or periodicity of the objective function $f$ \cite{Shahriari.2016}.
They are so-called positive-definite functions, often also called kernels \cite{Rasmussen.2008, Duvenaud.2014}.
It specifies the relationship between two 'points' (vector of the input space) $x$ and $x'$ and the corresponding changes in $f$ at these points. A covariance function is described by a set of parameters, also called hyperparameters, describing a specific behavior. This is how prior information is embedded in the Bayes optimization procedure.
Also in this work, the most commonly used covariance function, the \emph{Squared Exponential} (SE) kernel (often also referred to as Gaussian kernel) is used. \cite{Frazier.08.07.2018}.

\subsection{Problem definition and computational procedure}
The goal of the presented application example is to propose a concept and a numerical procedure for optimal robust seed train design, where robustness refers to the reproducibility of the seed train regarding viable cell density, meaning that despite initial uncertainties and variabilities concerning cell growth, low variations of viable cell density at the end of the seed train are obtained.

\begin{figure}[h]
	\centering
	\includegraphics[width=0.9\linewidth]{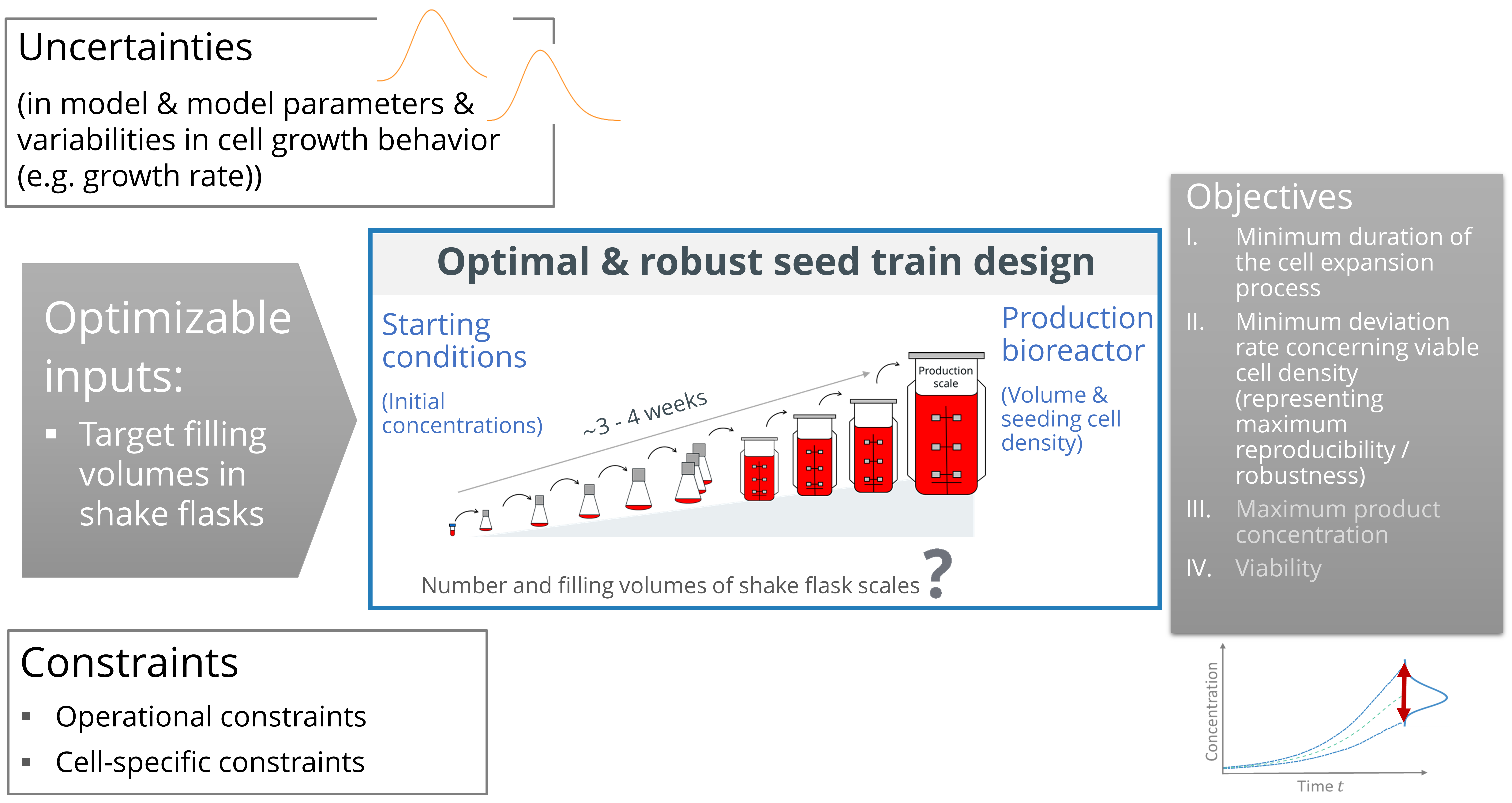}
	\caption{Goal of the study is to propose a concept and a numerical framework for optimal robust seed train design (blue box in the middle), including optimizable inputs (first gray box) as well as objectives (objective criteria) used in this study (right gray box).}
	\label{fig: motivation}
\end{figure}

First, seed train constraints are defined based on a chosen cell line and its characteristics concerning optimal cultivation conditions and based on the operative possibilities (e.g.\ feasible points in time for cell passaging).
Second, the optimizable input parameters and objective criteria (objective response variables) applied in this study are defined (as also illustrated in Figure~\ref{fig: motivation}), followed by the formulation of the mathematical optimization problem.
Thereafter, the optimization problem is solved using a workflow which connects seed train simulation and Bayes optimization. \\

The following objective criteria were chosen to represent an optimal seed train: I) a minimum duration ($d$) (=\,required cultivation time) of the seed train and II) a minimum deviation rate ($D$) regarding viable cell density, i.e.\ the probability that the seed train will run outside predefined ranges of viable cell density (for both, seeding viable cell density and transfer viable cell density)\footnote{This is important to consider because in case that specific constraints are not fulfilled, the performance of the cells could decrease. The growth rate could decrease and furthermore it has been observed that the violation of constraints could also cause less viability of the cells in the production phase \cite{Bohl.2020}.
} (compare to  Figure~\ref{fig: motivation} right gray box). These two attributes shall enable an optimal start of the production scale. Note that, in addition to these criteria, the growth rate is another important parameter affecting an optimal start of the production scale and the growth rate should be high until the end of the seed train. However, in this first optimization study it is not set as optimization criterion because the here defined seed train setup (in terms of medium concentrations and possible cultivation volumes per scale) together with the aim to reduce cultivation time already supports a good growth during the entire cultivation. But for other seed train setups it might be advisable to include growth rate at the end of the seed train into the optimization problem.\\
After consideration of the two mentioned objective criteria, a third and fourth objective criterion, the product concentration (titer) and the viability at the end of the cultivation in the production bioreactor (in this simulation study: after 8 days in batch mode, i.e.\ without addition of nutrient feeds) are added to the optimization problem (see Figure~\ref{fig: motivation} right gray box (III)). 
{\em Note: The authors are aware of the fact that cultivation in the production vessel itself, which is often performed in fed-batch mode, is also influenced by several process parameters having an impact on product quantity and quality. Moreover, data of further attributes would be necessary to describe product quality (e.g.\ of a recombinant therapeutic protein or antibody) but these are not provided and therewith not considered in this study}.\\
The input variables that can be varied to optimize the recently mentioned objective criteria, and thus the optimizable input variables, are the filling volumes in the first five shake flask scales, $V_1,...,V_5$ (compare to  Figure~\ref{fig: motivation}, the part of the seed train between thawing cells from a small vial and inoculation of the first biorector). These target values are important inputs of the seed train simulation process because they are used to calculate points in time for cell passaging. Volumes in the finally proposed seed train protocol (output of the seed train simulation) may vary within allowed working volume ranges and these are also presented in this work.

\paragraph{Formulation of the mathematical optimization problem} \, 
The optimizable variables and therewith inputs of the optimization problem are the filling volumes of the $n$ shake flask scales, $V_1,...,V_n$ which are included in the input vector 
\begin{align}
	x = (V_1,...,V_n)^T.
\end{align}

Outputs of the optimization problem are the defined objective criteria. These are seed train duration $d$ and deviation rate $D$ for the first optimization example. Thus, the unknown objective function (which should be minimized) can be written as follows:
\begin{align}
	f(x) = (f_1(x),f_2(x))^T
\end{align}
with $f_1(x)\,\hat{=}\,d$ and $f_2(x)\,\hat{=}\,D$.

The second optimization example includes a third and fourth optimization criterion, product concentration and viability at the end of the production scale (here after 8 days in the production vessel). Thus $f(x)$ expands to
\begin{align}
	f(x) = (f_1(x),f_2(x),f_3(x),f_4(x))^T
\end{align}
with $f_1(x)\,\hat{=}\,d$, $f_2(x)\,\hat{=}\,D$, $f_3(x)\,\hat{=}\,c_{\rm titer,end}$ and $f_4(x)\,\hat{=}\,\mbox{Viability}_{\rm end}$.

\subsection{Connecting seed train simulation and Bayes Optimization}
\begin{figure}[h]
	\centering
	\includegraphics[width=1\linewidth]{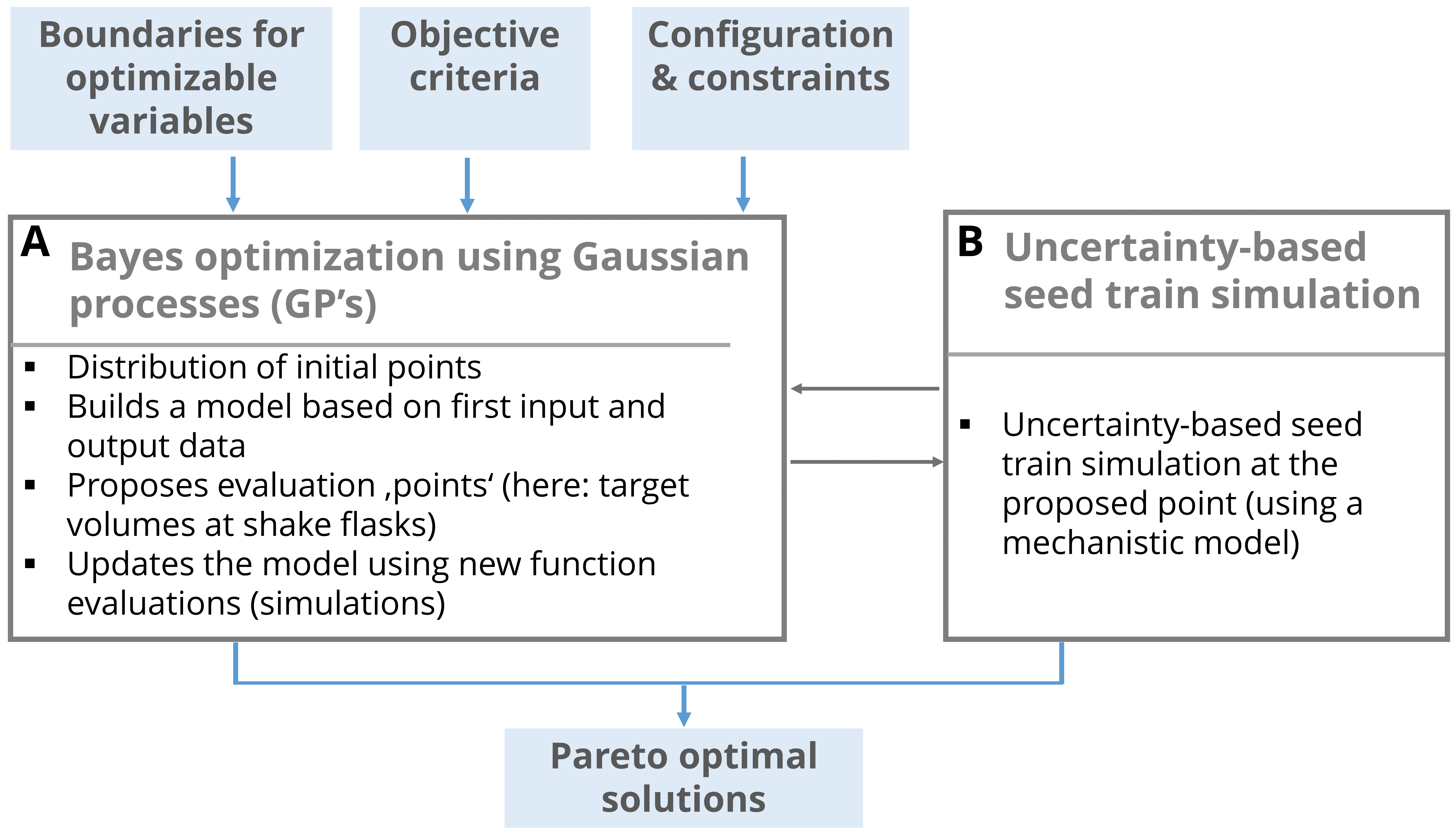}
	\caption{Scheme showing the applied computational workflow comprising {\bf A} a Bayes optimization algorithm which is coupled with {\bf B} a seed train simulation routine. Input and output values are shown in the blue boxes above and below.}
	\label{fig: scheme_bayes_opt}
\end{figure}

Uncertainty-based seed train simulation as described in Section~\ref{method: seed train simulation} was coupled with algorithms for Bayes optimization as described in Section ~\ref{method: Bayes_optimization}. The workflow integrating both components is illustrated in Figure~\ref{fig: scheme_bayes_opt}.
Input of the combined framework are the input variables: 
Boundaries for the optimizable variables (here filling volumes) and objective criteria (here seed train duration, deviation rate and in the second example also product concentration at the end of production scale) given all required seed train configuration settings and constraints (e.g.\ initial concentrations, practically feasible points in time for cell passaging, acceptable ranges for viable cell density, ...). 

First points (=\,combinations of optimizable variables) are determined using a Latin Hypercube design distributing these points within the design space (see Figure \ref{fig: scheme_bayes_opt}, Box A). Seed train simulations are performed at these points in order to obtain the corresponding objective criteria values. Input values together with output values form a data set. 
An unknown model describing the relationship between inputs and outputs is approximated through a Gaussian process (GP) which has to be trained (see Figure \ref{fig: scheme_bayes_opt}, Box A) based on the given data set.
Therefore, the Gaussian process proposes a point that has to be evaluated next (see Figure \ref{fig: scheme_bayes_opt}, Box B). 

A robust seed train is simulated, using a mechanistic process model, and the objective criteria are calculated. This output is then returned to the Bayes optimization (Box A) to update the GP.
Usually, experiments are performed to return the experimental output. The present approach instead exploits the advantages of the model-based upstream simulation in order to reduce the experimental effort to a minimum.

These steps are repeated various times, e.g.\ until a previously defined number of maximum iteration steps is reached. The latter depends on the resources (human and financial resources in case of laboratory experiments or computational resources in case of in-silico experiments).
In every iteration the Gaussian process chooses a new point aiming to move to the optimum and at the same time to reduce model uncertainty. 

Results of this optimization framework are the set of Pareto optimal setups (also called Pareto front) and their corresponding response values.

\subsection{Numerical solvers and tools}\label{subsec: system}
The programming language and numeric computing environment MATLAB \cite{MATLAB.2019} was used for the seed train simulations. The code for the optimization workflow was written in Python \cite{vanRossum.2010} using the MATLAB Engine API for Python to call MATLAB as a computational engine from Python code. To perform Bayes optimization within this workflow, the library GPflow \cite{Knudde.10.11.2017} was used.

\section{Results \& Discussion}

\subsection{Optimization of cultivation vessels regarding number of shake flask scales and filling volumes for 5, 4 and 3 shake flask scales}
In this section, it is investigated which cultivation filling volumes should be used for the flask scales in order to obtain optimal results in terms of seed train duration and deviation rate, here defined as the probability that the seed train will run outside the predefined acceptable ranges for initial viable cell density (VCD) and transfer VCD (final VCD before transfer into the next cultivation vessel) per scale. The latter is a measure for the robustness of the seed train regarding viable cell density.\\
For assessment of the optimization results, a conventional reference seed train comprising 5 shake flask scales was simulated based on a non-optimized design. Therefore, a common passaging interval of 3 days per cultivation scale was fixed and filling volumes were determined following a conservative layout (i.e.\ choosing not too huge differences between one cultivation scale and the next to ensure that enough viable cells are generated even if they grow a little bit slower than expected). \\
In a first step, the optimal combination of filling volumes for 5 shake flask scales is investigated and the results are compared to the reference seed train.
Afterwards, it is investigated if a reduction of shake flask scales from 5 to 4 or 3 shake flask scales leads to similar or even better results in terms of seed train duration and deviation rate. 
The number of bioreactor scales was kept fixed. Three bioreactors with filling volumes of 40 L, 320 L and 2100 L were used as pre-stages before inoculation of the production bioreactor with 9600 L.
The assumed seed train setup is given in Table \ref{tab: setup_bayes_opt_ex_2}.

\noindent To find the optimal solution, multi-objective Bayesian optimization coupled with uncertainty-based seed train simulation, as described in Section~\ref{method: seed train simulation}, was applied. 
First, a Latin hypercube design for $n_{\rm lhs}$ design 'points' (combinations of filling volumes, here $n_{\rm lhs}=10$) was initiated and seed train simulation was applied to calculate the objective criteria values, here deviation rate $D$ and seed train duration $d$ (replacing the normally required experimental cultivation runs) at each point. 
Within the Bayes optimization procedure, Gaussian processes were trained based on the simulation outcomes and an acquisition function was calculated in each iteration step in order to propose which point should be evaluated next.
The input space for shake flask filling volumes (here the optimizable variables) was defined as described in Table~\ref{tab: input_space}, assuming the possibility of using several shake flasks in parallel for one shake flask scale and also considering their working volumes ranges.

\begin{table}[h]
	\centering
	\renewcommand{\arraystretch}{1.3} 
	\caption[Input space]
	{Input space for the shake flask filling volumes, containing the possible filling volumes per scale, given for optimization runs with 5, 4 or 3 shake flask scales.}
	\label{tab: input_space}
	\begin{tabular}{l|lll}
		\hline
		& \multicolumn{3}{c}{Filling volumes}         \\
				& Range for   			& Range for             &  Range for   \\
				& 5 shake flask scale [L] 	& 4 shake flask scales [L]  &  3 shake flask scales  [L]   \\ 
		\hline
		V1		&  0.014 -- 0.015 & 0.014 -- 0.015   & 0.014 -- 0.015     \\
		V2		&  0.05 -- 0.15  & 0.1 -- 1         & 0.1 -- 2  \\
		V3		&  0.15 -- 1.5   & 1.5 -- 4         & 4 -- 8    \\
		V4      &  1.5 -- 4       & 4 -- 8           & -    \\
	    V5      &  4 -- 8         & --               & -    \\
		\hline
	\end{tabular}
\end{table}

\subsubsection{Optimization of 5 shake flask scales}\label{res: 5_sf} 
\begin{figure}[h]
	\centering  
	\includegraphics[width=0.7\linewidth]{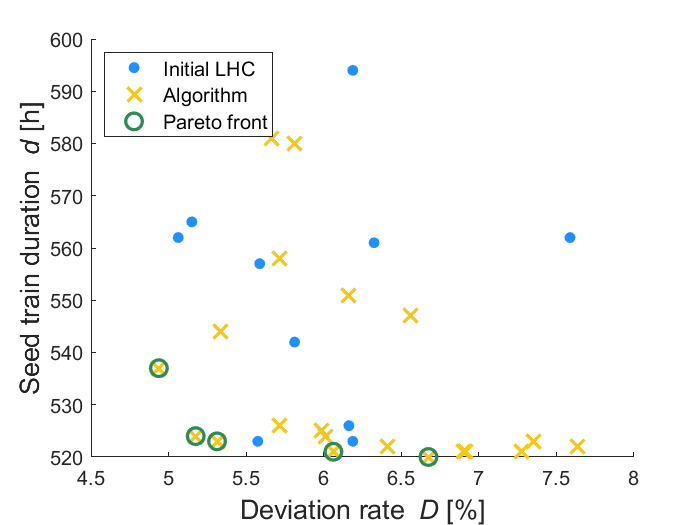}
	\caption[Pareto front showing the objective criterion seed train duration over the objective criterion deviation rate; optimizable variables are the filling volumes of the 5 shake flask scales (blue dots: based on the initial Latin hypercube design; yellow crosses: based on the proposed points; green circles: Pareto optimal solutions).]{Objective criterion {\em seed train duration} over objective criterion {\em deviation rate} obtained for evaluated points (optimizable variables, here combinations of 5 shake flask filling volumes); blue dots: based on the initial Latin hypercube design (LHC); yellow crosses: based on the points proposed by the algorithm; green circles: Pareto optimal solutions (=\,Pareto front)).}
	\label{fig: pareto_front_sf5_mu1}
\end{figure}

\, \\
The first optimization was performed for a seed train comprising 5 shake flask scales.
Figure~\ref{fig: pareto_front_sf5_mu1} shows the objective criteria values for each evaluated point, whereby the outcomes based on the initial Latin hypercube space are illustrated by blue dots and the outcomes for the proposed points based on the trained Gaussian processes are illustrated through yellow crosses. The optimal solutions are those near to the lower left corner aiming to minimize seed train duration and the deviation rate. The Pareto optimal solutions, also called non-dominated solutions are illustrated through green circles. 
A solution (here seed train setup/combination of filling volumes) is called non-dominated if no solution exists leading to better (here lower) objective criteria values. As described previously, several Pareto optimal solutions can be obtained because when considering two or more objective criteria then for two different solutions one criterion might have better (here lower) value then the other solution for the same objective, while the other criterion has worse (here higher) values. The set of all Pareto optimal solutions is called Pareto-front.

\noindent For the investigated scenario (5 shake flask scales and the seed train configuration according to Table~\ref{tab: setup_bayes_opt_ex_2}) five Pareto optimal solutions were obtained (see green circles in Figure~\ref{fig: pareto_front_sf5_mu1}). It can be seen that comparing two of these solutions (green circles) each, one solution has a lower (here better) seed train duration value than the other solution and the opposite holds for the deviation rate.\\
The corresponding values for the optimizable variables, here shake flask filling volumes ($V_1$, $V_2$, $V_3$, $V_4$ and $V_5$), and the corresponding objective criteria values, here deviation rate $D$ and seed train duration $d$, are listed in Table~\ref{tab: bayes_opt_5sf_volumes}.

\begin{table}[h]
	\centering
	\renewcommand{\arraystretch}{1.3} 
	\caption[Pareto optimal solutions concerning the choice of filling volumes in shake flask scales, for three scenarios for 5 flask scales. The following bioreactor filling volumes are 40\,L, 320\,L and 2210\,L. The averaged filling volumes in L and the resulting deviation rate ($D$) in \% and seed train duration $d$ in h are listed for each solution.]
	{Pareto optimal solutions concerning the choice of filling volumes in shake flask scales, for three scenarios for 5 flask scales. The following bioreactor filling volumes are 40\,L, 320\,L and 2210\,L. The averaged filling volumes in L and the resulting deviation rate ($D$) in \% and seed train duration ($d$) in h are listed for each solution.}
	\label{tab: bayes_opt_5sf_volumes}
	\begin{tabular}{l|lllll|ll}
		\hline
		& \multicolumn{5}{c}{Filling volumes}  &    &    \\
		Solution & Vol. 1 & Vol. 2 &  Vol. 3 & Vol. 4 & Vol. 5 & D   &  d  \\
		& [L] 	  & [L]	   & [L]     & [L]    & [L]    & [\%] &  [h]   \\ 
		\hline
		1		&  0.015  & 0.065   & 0.904    & 2.355  & 7.78 & 537  & 4.9  \\
		2		&  0.015  & 0.115   & 0.451    & 1.672  & 7.89 & 521  & 6.1  \\
		3		&  0.015  & 0.104   & 0.340    & 1.614  & 6.85 & 524  & 5.2  \\
		4       &  0.014  & 0.103   & 0.369    & 1.582  & 7.87 & 523  & 5.3  \\
		5       &  0.015 & 0.114   & 0.431    & 2.026  & 7.97 & 520  & 6.7  \\
		\hline
		 & \multicolumn{5}{c}{Filling volumes} of reference seed train  & & \\
		Reference		& 0.015  & 0.08 & 0.30 & 2 & 4  & 41.7 & 576 \\ 
	\end{tabular}
\end{table}

\noindent The filling volume of the first scale was limited to a very narrow range (14--15 mL) (A higher variation after cell thawing  was not expected).
Most obtained solutions start with the maximum value of this range (see Table~\ref{tab: bayes_opt_5sf_volumes}, first column).
The filling volume of flask scale 2 varies between 0.065 and 0.115 L, the filling volume of flask scale 3 between 0.340 and 0.904~L, the filling volume of flask scale 4 between 1.582 and 2.355 L and of flask scale 5 between 6.85 and 7.97~L. All five combinations lead to a deviation rate $D$ of less than 7\% and to a seed train duration between 520 to 537~h. 

\begin{figure}
	\centering
	\includegraphics[width=1\linewidth]{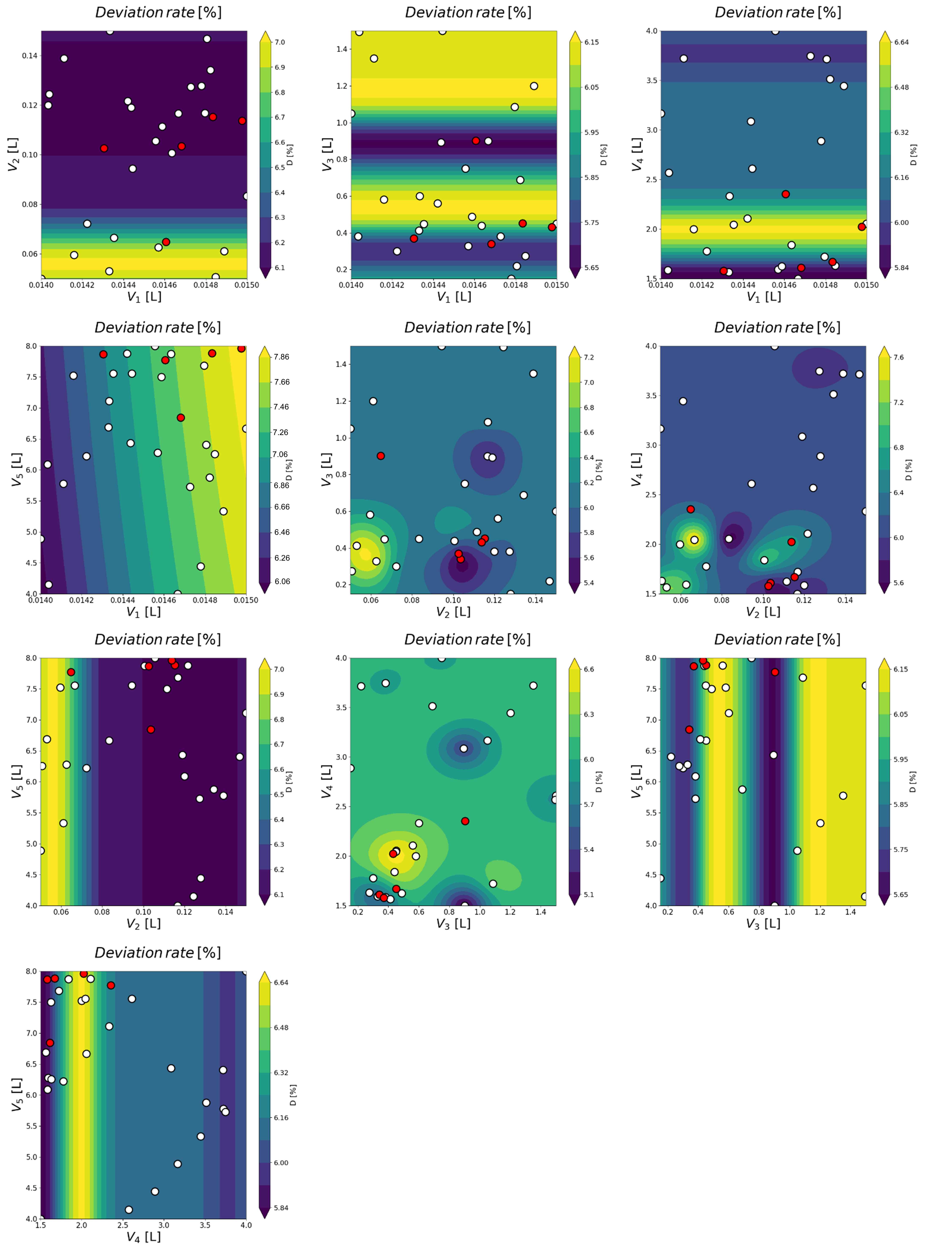}
	\caption{Contour plots showing two optimizable variables on x and y-axis and one objective (here Deviation rate $D$ in \%), assigned to each combination of the two variables, through colored isolines. For example, the diagram top left shows the deviation rate for each combination of $V_1$ (filling volume in flask scale 1) and $V_2$ (filling volume in flask scale 2) through colors representing the corresponding values in \%, as indicated on the color bar. Moreover, the results obtained through seed train simulations are shown by dots. The red dots represent the non-dominated (optimal solutions).}
	\label{fig: contour_plots_D_5sf_mu1}
\end{figure}

\begin{figure}
	\centering
	\includegraphics[width=1\linewidth]{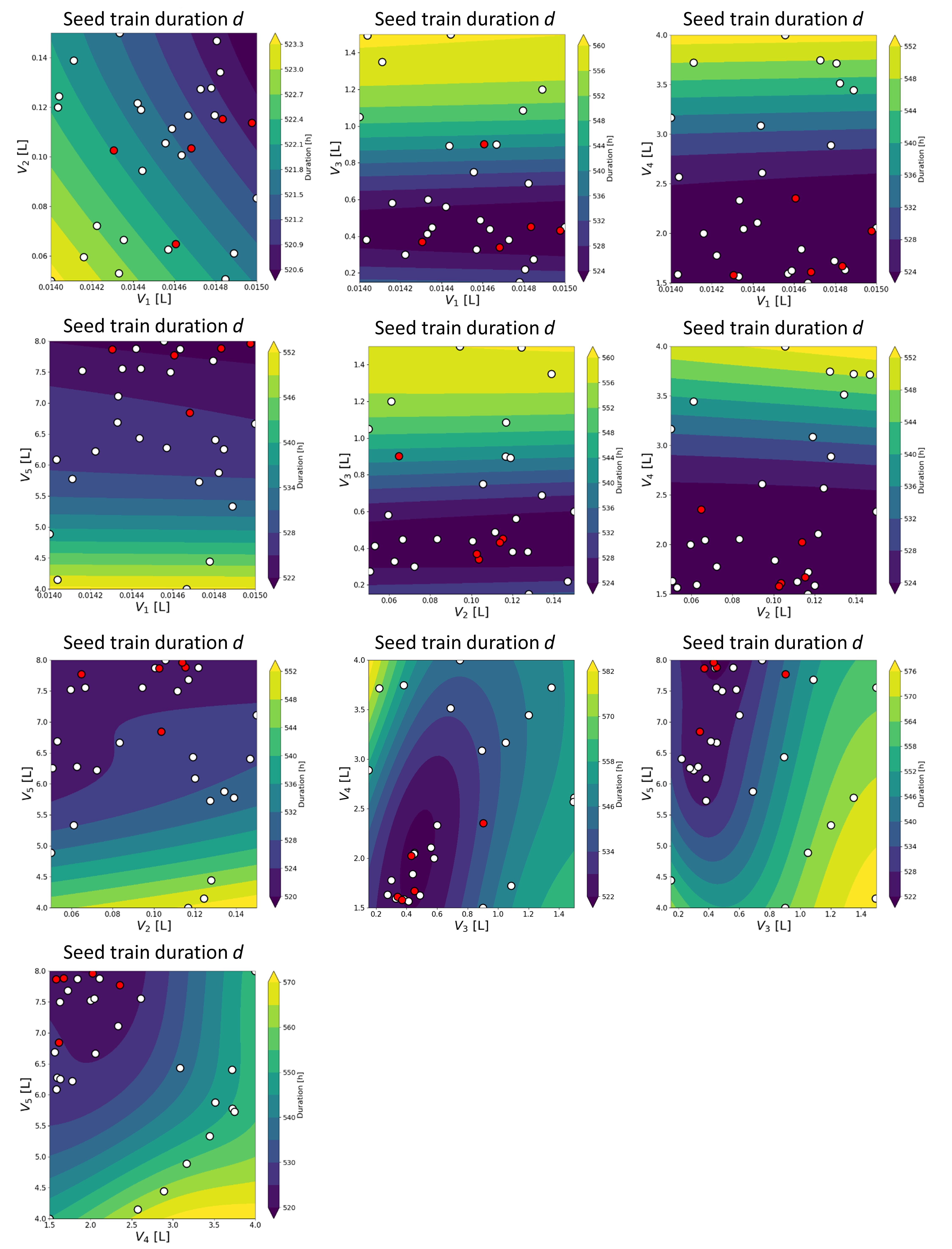}
	\caption{Contour plots showing two optimizable variables on x and y-axis and one objective (here seed train duration ($d$) in h), assigned to each combination of the two variables, through colored isolines. For example, the diagram top left shows the deviation rate for each combination of $V_1$ (filling volume in flask scale 1) and $V_2$ (filling volume in flask scale 2) through colors representing the corresponding values in \%, as indicated on the color bar. Moreover, the results obtained through seed train simulations are shown by dots. The red dots represent the non-dominated (optimal solutions)}
	\label{fig: contour_plots_duration_5sf_mu1}
\end{figure}

A more detailed illustration of the obtained results is presented in Figure~\ref{fig: contour_plots_D_5sf_mu1} and Figure~\ref{fig: contour_plots_duration_5sf_mu1}. For two optimizable variables and one objective criterion each (deviation rate in Figure~\ref{fig: contour_plots_D_5sf_mu1} and seed train duration in Figure~\ref{fig: contour_plots_duration_5sf_mu1}), a contour plot is shown which illustrates the objective value for each calculated point (combination of the two variables), using the trained Gaussian processes, through colored isolines. \\ 
For example, the diagram in the top left of Figure~\ref{fig: contour_plots_D_5sf_mu1} shows the deviation rate for each combination of $V_1$ (filling volume in flask scale~1) and $V_2$ (filling volume in flask scale~2) through colors representing the corresponding values in \%, as indicated in the color bar. The results obtained through seed train simulations are shown by dots. The red dots represent the non-dominated (optimal solutions), optimal with respect to the defined multi-objective optimization problem. The dark blue area indicates combinations of $V_1$ and $V_2$ leading to a lower deviation rate. It can be seen that values above 0.1 for $V_2$ combined with any value of $V_1$ (within the given range) lead to the lowest deviation rates (below 6.2\%, see dark blue area).
Also the optimal solutions (red dots) are mostly located in the area with higher filling volumes for shake flask 2, $V_2$, except one (red dot at $V_2\approx$\,0.065). 

For some combinations, a closer delimitation is possible. For example, the middle diagram in the second row ($V_3$ over $V_2$) shows a limited region (dark blue area) and therewith a specific combination of $V_3$ and $V_2$ that leads to the lowest deviation rates (\textless~5.6\%)). These are around 0.3~L for $V_3$ and around 0.105~L for $V_2$.

Analogously, Figure~\ref{fig: contour_plots_duration_5sf_mu1} shows the contour plots for the second objective criterion, seed train duration. The dark blue areas show the combinations with the lowest seed train durations (approximately below 528\,h).
It can be seen in these diagrams that most red dots are located in the dark blue regions. 
For some combinations the dark blue areas are wider, distributed over several possible values for one variable, e.g.\ the diagram in the top center, top left, center, and center right. \\
Other combinations show narrower regions with low seed train durations as can be seen in the diagram showing $V_4$ over $V_3$. 
The lowest seed train duration is obtained for filling volumes between 1.5 and 2.5~L for shake flask 4 in combination with filling volumes between 0.2 and 0.8~L for shake flask 3. \\
Overall, these diagrams give an overview of the impact of two combined optimizable variables each on a specific objective criterion.

\begin{figure}[h]
	\centering
	\includegraphics[width=1\linewidth]{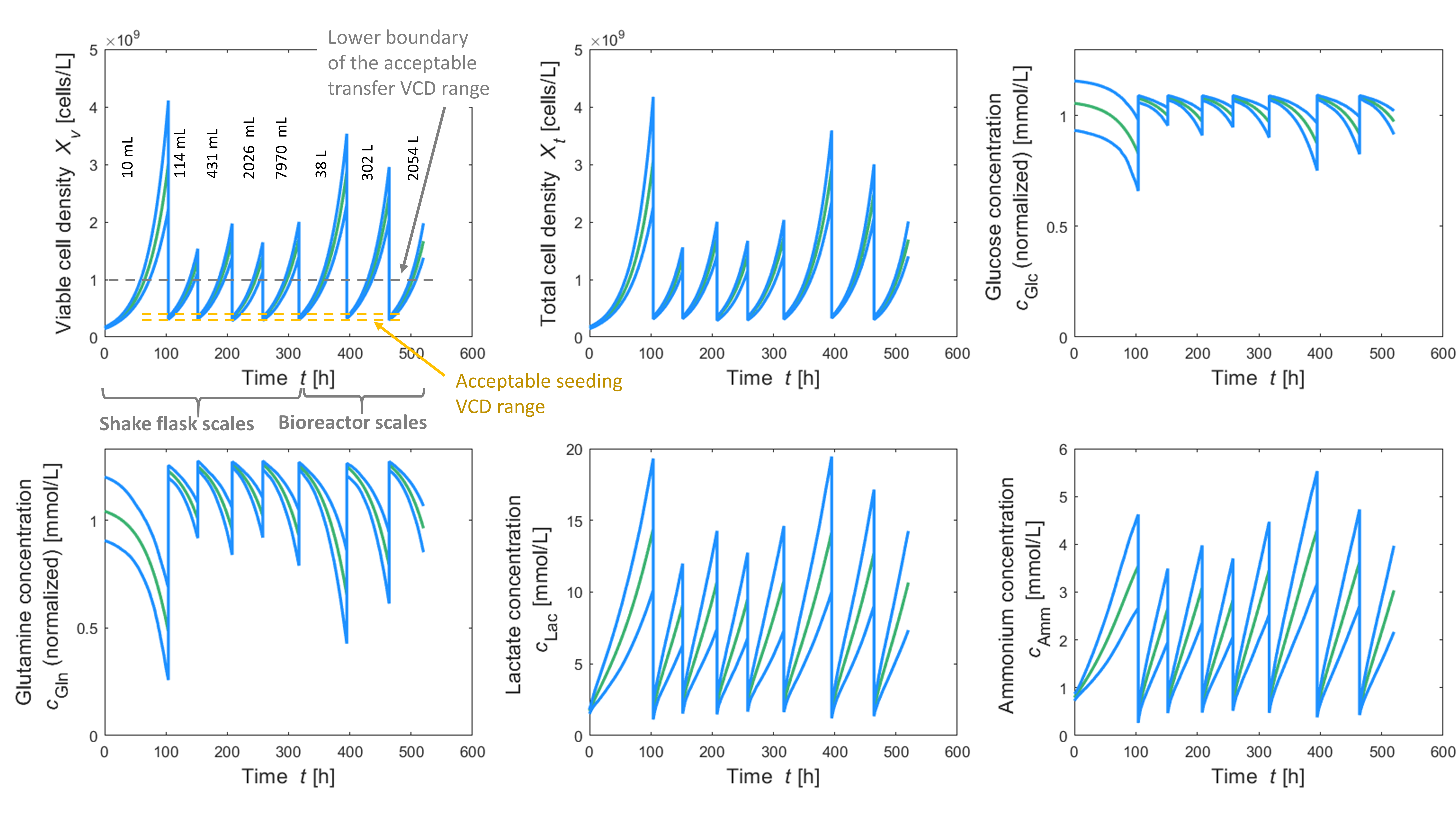}
	\caption{Seed train showing viable cell density (VCD) and total cell density as well as substrate (glucose and glutamine) and metabolite (lactate and ammonium) concentrations over time and over the whole seed train (5 shake flask scales and three bioreactor scales), based on the shake flask filling volumes according to solution 1. The green lines represent the mean time course and the blue lines show the corresponding 90\%-prediction band (5\%- and 95\%-quantiles).
	The plot (top left) also inlcudes the filling volumes and the acceptable ranges for seeding VCD and transfer VCD, illustrated through dashed lines.}
	\label{fig: seed train sol1 5sf}
\end{figure}

In addition to this information, simulated time profiles (predictive mean in green, 90\% prediction bands in blue) of viable and total cell density as well as concentrations of glucose, glutamine, lactate and ammonium (see Figure~\ref{fig: seed train sol1 5sf}) can be obtained for each solution as well as a seed train protocol containing information about the calculated passaging intervals, amount of medium, etc.
It can be seen in Figure~\ref{fig: seed train sol1 5sf} top left that based on the given filling volumes in addition to the flexibility to choose individual points in time for cell passaging in each scale, it is possible to set the seeding viable cell density at the beginning of each cultivation scale on the desired value with low variability, allowing to stay within the corresponding acceptable ranges for seeding VCD (see yellow dashed lines). Also transfer VCDs lie within the corresponding acceptable range with high probability (see lower boundary, gray dashed line). Moreover, it can be seen that substrate concentrations are not depleted and according to \cite{HernandezRodriguez.2019}, values of 20 mmol/L lactate and 5 mmol/L ammonium are not yet inhibiting concentrations for this cell line. 

\begin{figure}[h]
	\centering
	\includegraphics[width=1\linewidth]{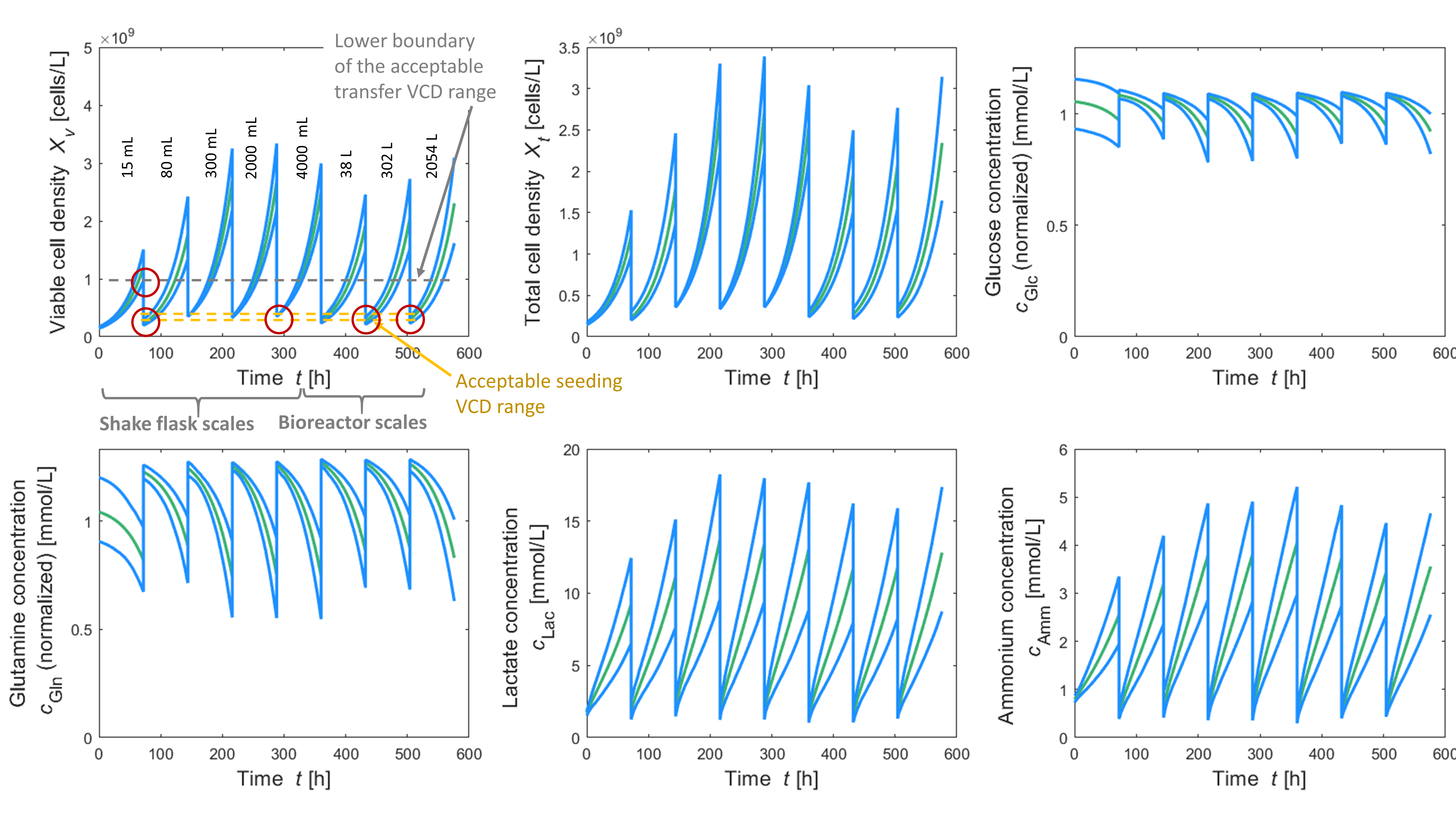}
	\caption{Reference (non-optimized) seed train showing viable and total cell density as well as substrate (glucose and glutamine) and metabolite (lactate and ammonium) concentrations over time, based on a reference configuration setup for 5 shake flask scales using passaging intervals of 72\,h each. The green lines represent the mean time course and the blue lines show the corresponding 90\%-prediction band (5\%- and 95\%-quantiles).}
	\label{fig: ref_seed train 5sf}
\end{figure}

For a better assessment, the obtained results are compared to the reference seed train which is also defined in this work for 5 shake flask scales and illustrated in Figure~\ref{fig: ref_seed train 5sf}. It grounds on a (non-optimized) configuration setup  for 5 shake flask scales using fixed passaging intervals of 72\,h each (common practice) and filling volumes of 15\,mL (flask scale 1), 80\,mL (flask scale 2), 300\,mL (flask scale 3), 2,000\,mL (flask scale 4) and 4,000\,mL (flask scale 5). This choice grounds on a rather conservative approach aiming to avoid the risk of reaching too low transfer cell densities at the end of a cultivation scale but without inclusion of probabilistic simulations.   \\
The proposed method instead includes risk calculations and a passaging strategy aiming to minimize this risk but at the same time identifying a seed train configuration which is optimal regarding further objectives like seed train duration in the present case.

A comparison of the seed train solutions obtained after optimization and the reference seed train shows that deviation rate is much lower after optimization (4.9--6.7\% instead of 41.7\%) and seed train duration could be reduced by 56 h from 576 h to 520 h.
Figure~\ref{fig: ref_seed train 5sf}, diagram top left shows where seeding or transfer viable cell density  do not lie fully within the acceptable ranges (see red circles). This is different for the optimized solutions, e.g solution 5, as illustrated in Figure~\ref{fig: seed train sol1 5sf}, where seeding VCD lies within the acceptable range and also transfer VCD lies above the lower bound of the acceptable range for transfer VCD.
This significant reduction in time ($\approx$\,2 days per seed train) would contribute to a meaningful acceleration of the production process. 

\subsubsection{Optimization of 3 and 4 shake flask scales}\label{res: 3_4_sf} \, \\
In a next step, the number of shake flask scales was reduced from 5 to 4 and then to 3 shake flask scales and the same optimization procedure was applied. The aim was to investigate if less cultivation vessels would lead to comparable results and if so, which target and filling volumes should be chosen. This is of interest because less operations (like transferring cells from one scale into another one) signify less risk of failure and deviations. 

\begin{figure}[h]
	\begin{minipage}{0.5\textwidth}
		\includegraphics[width=1\linewidth]{"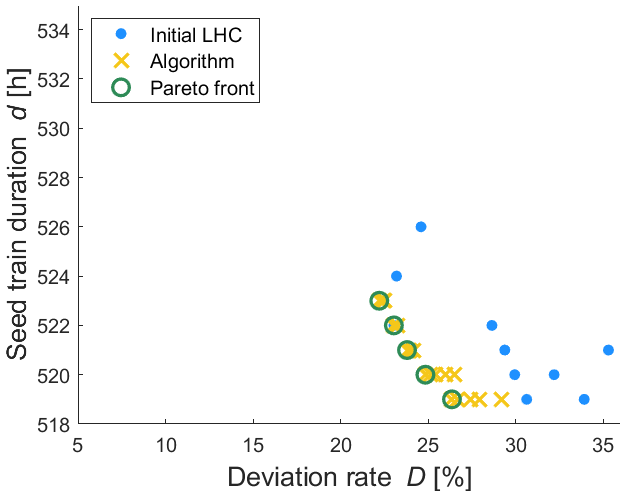"}
	\end{minipage}
	\begin{minipage}{0.5\textwidth}
 		\includegraphics[width=1\linewidth]{"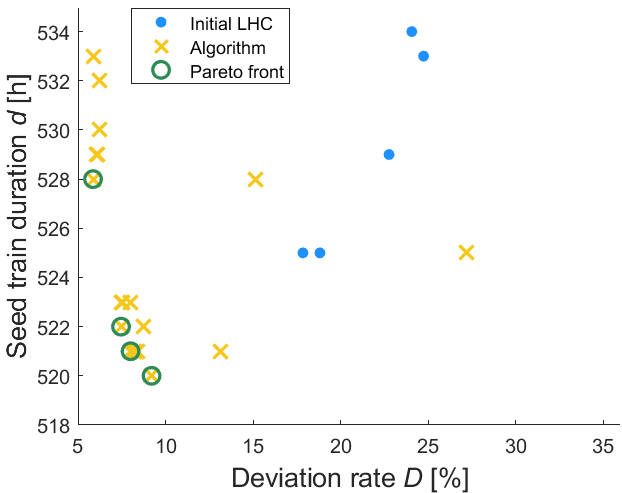"}
	\end{minipage}
    \caption[Pareto optimal solutions regarding seed train duration and deviation rate for 3 and 4 shake flask scales]{Objective criterion {\em seed train duration} over objective criterion {\em deviation rate} obtained for evaluated points (optimizable variables), here combinations of for 3 (left) and 4 shake flask scales (right); blue dots: based on the initial Latin hypercube design (LHC); yellow crosses: based on the points proposed by the algorithm; green circles: Pareto optimal solutions (=\,Pareto front).}
    \label{fig: bayes_opt_3_4_sf}
\end{figure}

Figure~\ref{fig: bayes_opt_3_4_sf} shows the obtained values for the objective criteria deviation rate and seed train duration for different combinations of filling volumes for 3 (left) and for 4 shake flask scales (right). Also here, the solutions based on the initial Latin hypercube design are shown by blue dots and Pareto optimal solutions are highlighted through green circles. 

It can be seen that for both scenarios, combinations of filling volumes could be found leading to an overall seed train cultivation time between 519 and 530~h. However, the scenario of using 4 shake flask scales, leads to lower deviation rates ($D\,<\,10\%$) compared to the scenario of using 3 shake flask scales ($23\% < D < 26\%$).

\noindent The corresponding filling volumes and the obtained filling volumes (based on the underlying passaging strategy) of the Pareto optimal solutions are listed in Table~\ref{tab: bayes_opt_3_4sf_volumes} together with the results for 5 shake flask scales from Table~\ref{tab: bayes_opt_5sf_volumes}. The results are sorted as discovered by the optimization algorithm. The obtained filling volumes for 4 shake flask scales are very similar, except for shake flask scale 4 ( $V_1$ = 15\,mL, $V_2$ = 158--200\,mL, $V_3$\,=\,1.51--1.60\,L and $V_4$\,=\,4.81--7.58\,L). Some of the obtained solutions would be seen or treated as equal in practice, because the differences are rather small. For example it would not be distinguished between 0.190 and 0.195~L. Probably 200~mL would be used instead. 
However, the applied optimization algorithm works on a continuous input space and differentiates between the solutions listed in the Table~\ref{tab: bayes_opt_3_4sf_volumes}, even though the differences are very low. 
The obtained optimal filling volumes for 3 shake flask scales also look similar, but with a bit more variation for shake flask 3 (4.45--5.58\,L).

Comparing the results for the three scenarios (3, 4 and 5 shake flask scales) endorse a decision against the 3 flask scales-scenario due to the higher deviation rates (>\,20\%), which stands for less process robustness. Between the other two scenarios (4 or 5 shake flask scales) only little differences with respect to deviation rates are observed for the determined optimal solutions (4.9\%--6.7\% for 5 shake flasks, 5.9\%--9.2\% for 4 shake flasks). Using 5 shake flask scales would lead to more or less similar 
cultivations times (520--537~h) but one operational step more would be required. 

This information together with the corresponding seed train protocol provides a solid basis to take a decision for one of the proposed optimal seed trains designs, taking into account seed train duration, robustness (expressed through deviation rates) and operational steps.

\begin{table}[h]
	\centering
	\renewcommand{\arraystretch}{1.3} 
	\caption[Pareto optimal solutions concerning the choice of filling volumes in shake flask scales, for three scenarios for 5 flask scales. The following bioreactor filling volumes are 40\,L, 320\,L and 2210\,L. The averaged filling volumes in L and the resulting deviation rate ($D$) in \% and seed train duration $d$ in h are listed for each solution.]
	{Pareto optimal solutions concerning the choice of filling volumes in shake flask scales, for 3, 4 and 5 shake flask scales. The following bioreactor filling volumes are 40\,L, 320\,L and 2210\,L. The averaged filling volumes in L, the resulting deviation rate ($D$) in \% and seed train duration $d$ in h are listed for each solution.}
	\label{tab: bayes_opt_3_4sf_volumes}
	\begin{tabular}{l|lllll|ll}
		\hline
		& \multicolumn{4}{c}{Filling volumes}  &    & &\\
		Solution  & Vol. 1 & Vol. 2   &  Vol. 3 & Vol. 4 & Vol. 5 & D   &  d   \\
		& [L] 	  & [L]	   & [L]     & [L]      &   [L]     & [\%]  &  [h]    \\ 
		\hline
			& \multicolumn{3}{l}{5 flask scales} & & & &\\
		1		&  0.015  & 0.065   & 0.904    & 2.355  & 7.78   & 4.9 & 537 \\
		2		&  0.015  & 0.115   & 0.451    & 1.672  & 7.89   & 6.1 & 521 \\
		3		&  0.015  & 0.104   & 0.340    & 1.614  & 6.85   & 5.2 & 524 \\
		4       &  0.014  & 0.103   & 0.369    & 1.582  & 7.87   & 5.3 & 523  \\
		5       &  0.0015 & 0.114   & 0.431    & 2.026  & 7.97   & 6.7 & 520 \\	
		\hline	
		& \multicolumn{3}{l}{4 flask scales} & & & &\\
		1		&  0.015  & 0.195   & 1.60    & 7.58 &  & 9.2    &  520   \\
		2		&  0.015  & 0.190   & 1.51    & 5.39 &  & 8.0    &  521   \\
		3		&  0.015  & 0.169   & 1.52    & 6.33 &  & 7.5    &  522   \\
		4		&  0.015  & 0.158   & 1.59    & 4.81 &  & 5.9    &  528   \\
		\hline
		& \multicolumn{3}{l}{3 flask scales} & & & &\\
		1 	& 0.015 & 0.733 & 4.45  &  & & 23.0 & 522 \\
		2 	& 0.015 & 1.046 & 4.85  &  & & 23.8 & 521 \\
		3 	& 0.015 & 1.103 & 5.26  &  & & 24.8 & 520 \\
		4 	& 0.015 & 0.934 & 4.77  &  & & 23.0 & 522 \\
		5 	& 0.015 & 1.110 & 4.65  &  & & 22.2 & 523 \\
		6 	& 0.015 & 1.306 & 5.58  &  & & 26.4 & 519 \\
		\hline
	\end{tabular}
\end{table}

\clearpage

\subsection{Application to further cell lines with potentially different growth rates}\label{res: variation_of_mu}
The optimization examples presented in the previous subsection were applied to a specific CHO cell line with growth characteristics described by a set of model parameters derived from an industrial cell culture process which was investigated in \cite{HernandezRodriguez.2019}.
If a different cell line or a clonal cell population with potentially differing growth behavior is used, then the optimization has to be performed for this specific cell line. In the following simulation study, a cell line having a 5\% lower and a cell line having a 5\% higher maximum cell-specific growth rate compared to the reference maximum growth rate ($\mu_{\rm max}$\,=\,0.028\,h$^{-1}$ for the first bioreactor scale and $\mu_{\rm max}$\,=\,0.029\,h$^{-1}$ for the remaining seed train scales) are assumed and the optimization is applied for both scenarios. \\
The results for the obtained/proposed filling volumes as well as the corresponding seed train duration and deviation rate are listed in Table~\ref{tab: bayes_opt_different_growth_rates}.

As expected, cells which grow faster (higher maximum growth rate $\mu_{\rm max}$) would require less time until reaching a specific target cell density. This can be seen in the right column of Table~\ref{tab: bayes_opt_different_growth_rates}. Using 5 flask scales, the optimal required seed train duration would lie between 494 and 503 h for a cell line with a 5\% higher growth rate compared to the reference cell line which would need 520--537\,h (see Table~\ref{tab: bayes_opt_5sf_volumes}). Correspondingly, cells with a 5\% lower growth rate would need more time (550--568 h). The same is observed when using 4 or 3 shake flasks. \\
With respect to the deviation rates which represent the robustness of the seed train design regarding variability of viable cells, it can be seen that low deviation rates of between 4.1\% and 11.6\% can be reached when using 5 or 4 flask scales, even if the maximum growth rate varies $\pm$ 5\%.
A critical limit was identified for the combination of using 3 shake flask scales for a slower growing cell line. The corresponding optimal solution shows a comparatively higher deviation rate (19.2--29.1\%) together with a high seed train duration (548--552\,h).
 
\begin{table}[h]
	\centering
	\renewcommand{\arraystretch}{1.3} 
	\caption[]
	{Pareto optimal solutions concerning the choice of filling volumes in shake flask scales, for 3, 4 and 5 shake flask scales, for two different scenarios. Scenario 1 assumes a 5\% lower and scenario 2 a 5\% higher cell-specific maximum growth rate compared to the reference maximum growth rate. The bioreactor filling volumes which follow after the shake flask scales are 40\,L, 320\,L and 2210\,L. The filling volumes in L, the resulting deviation rate ($D$) in \% and seed train duration ($d$) in h are listed for each solution (several Pareto optimal solutions can be obtained per setup).}
	\small
	\label{tab: bayes_opt_different_growth_rates}
	\begin{tabular}{l|lllll|ll}
		\hline
		& \multicolumn{5}{c}{Filling volumes}  &    &        \\
		Solution & Vol. 1 & Vol. 2 &  Vol. 3 & Vol. 4 & Vol. 5 & D   &  d  \\
		& [L] 	  & [L]	   & [L]     & [L]    & [L]    & [\%] &  [h]   \\ 
		\hline
		{\bf 5 flask scales} & \multicolumn{3}{l}{5 flask scales} & & & &  \\
		5\% lower growth rate & & & & & & &  \\
		1 	&  0.015  & 0.083    & 0.45    & 2.06  & 6.67  & 7.3 & 550   \\
		2	&  0.014  & 0.072    & 0.30    & 2.78  & 6.23  & 7.0 & 551   \\
		3 	&  0.014  & 0.105    & 0.45    & 2.53  & 6.77  & 6.1 & 553   \\
		4 	&  0.015  & 0.119    & 0.84    & 2.96  & 6.21  & 5.5 & 568   \\
		5 	&  0.014  & 0.122    & 0.56    & 2.64  & 6.62  & 5.8 & 559   \\
		5\% higher growth rate & & & & & & &  \\
		6 	&  0.015  & 0.08    & 0.354    & 1.52   & 7.24  & 6.2 & 495   \\
		7 	&  0.014  & 0.09    & 0.560    & 1.89   & 7.19  & 4.6 & 502   \\
		8 	&  0.014  & 0.14    & 0.545    & 1.89   & 7.16  & 4.1 & 503   \\
		9 	&  0.015  & 0.06    & 0.313    & 1.63   & 7.72  & 6.5 & 494   \\
		10 	&  0.014  & 0.09    & 0.312    & 1.82   & 7.08  & 5.3 & 496   \\
		11 	&  0.014  & 0.13    & 0.564    & 2.16   & 7.41  & 4.8 & 501   \\		
		\hline
	    {\bf 4 flask scales} & \multicolumn{3}{l}{4 flask scales}  & & & &  \\
		5\% lower growth rate & & & & & & &  \\
		12 & 0.015 & 0.158 & 1.99 & 7.6 & & 8.8  & 551 \\
		13 & 0.015 & 0.147 & 1.59 & 7.7 & & 7.7  & 552 \\
		14 & 0.015 & 0.132 & 2.00 & 7.8 & & 7.6  & 553 \\
		15 & 0.015 & 0.167 & 1.56 & 7.5 & & 9.8  & 550 \\
		16 & 0.015 & 0.180 & 1.58 & 7.4 & & 11.6 & 548 \\	
		5\% higher growth rate & & & & & & &  \\
		17  & 0.015 & 0.199 & 1.59 & 5.2 & & 5.3  & 501 \\
		18 	& 0.015 & 0.246 & 1.53 & 7.9 & & 10.5 & 493 \\
		19 	& 0.015 & 0.215 & 1.53 & 7.2 & & 7.7  & 494 \\
		20 	& 0.015 & 0.193 & 1.56 & 6.8 & & 6.8  & 496 \\
		21 	& 0.015 & 0.191 & 1.60 & 5.6 & & 5.8  & 498 \\
		22 	& 0.015 & 0.210 & 1.56 & 7.5 & & 7.3  & 495 \\
		23 	& 0.014 & 0.183 & 1.71 & 6.1 & & 5.4  & 499 \\
		\hline
	     {\bf 3 flask scales} & \multicolumn{3}{l}{3 flask scales} & & & &  \\
		5\% lower growth rate & & & & & & &  \\
		24 	& 0.015 & 0.174 & 4.14  &  & & 20.0 & 550 \\
		25 	& 0.015 & 1.011 & 4.03  &  & & 25.7 & 549 \\
		26 	& 0.015 & 0.151 & 4.02  &  & & 19.2 & 552 \\
		27 	& 0.015 & 0.929 & 4.34  &  & & 29.1 & 548 \\
		5\% higher growth rate & & & & & & &  \\
		28 	& 0.015 & 0.235 & 4.277  &  & & 11.13 & 496 \\
		29 	& 0.015 & 0.987 & 7.683  &  & & 25.5  & 493 \\
		30 	& 0.015 & 0.267 & 4.589  &  & & 14.9  & 495 \\
		\hline
	\end{tabular}
\end{table}


\subsection{Optimization regarding four objectives including product concentration}\label{res: 3_objectives}
To show the applicability of the proposed method to more than two objectives, a third and a fourth objective criterion, titer concentration and viability at the end of the production vessel (after 8 days) was added.
Whereas the first two objective criteria (seed train duration and deviation rate) are related to the seed train itself, the third and fourth criterion refer to the generated product in the production vessel and to the viability of the cells in the production vessel. Product concentration as well as product quality can be influenced by many factors (seeding cell density, substrate concentrations and nutrient feeds, metabolite production, temperature, pH, dissolved oxygen and carbon dioxide concentration, osmolality and more) and also by the amount and the state of the cells at the end of the seed train. Since no data describing product quality are available, product concentration and viability are considered in this study. A further simplification that was made is the assumption that the production vessel is performed in batch-mode (meaning without any addition of nutrient feeds or medium renewals). The reason for this simplification is to avoid confounding effects. 
The authors are aware of the fact that many factors affect product concentration and product quality and when data of other critical process parameters or quality attributes are available, these could also be considered in the same manner. The main purpose of the present simulation example is to demonstrate how the proposed method can be applied to more than two objectives and how the corresponding results can be illustrated and interpreted.

\begin{figure}[h]
	\centering
	\includegraphics[width=0.8\linewidth]{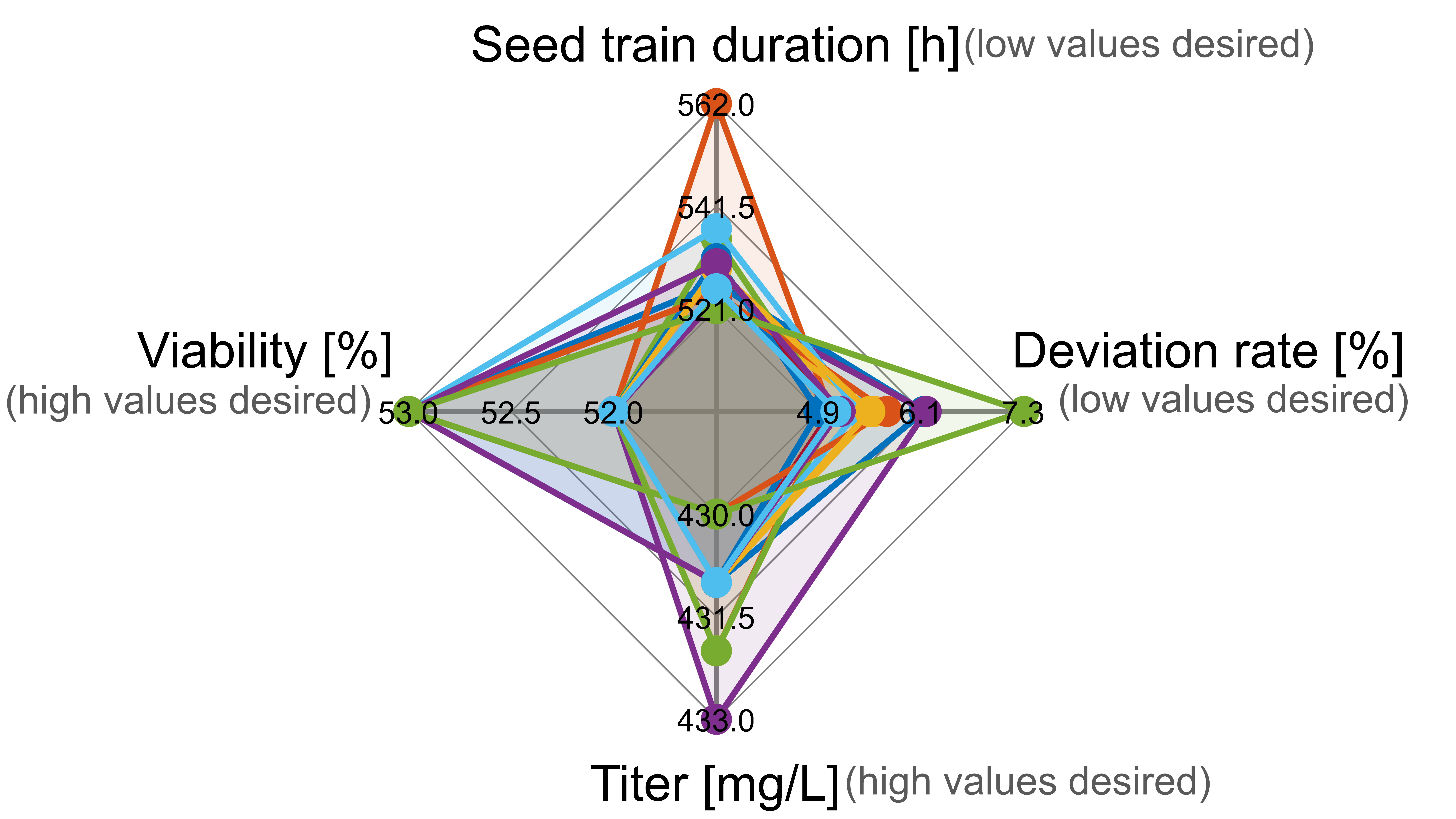}
	\caption{Spider plot showing the objective criteria values (seed train duration, deviation rate, titer and viability after 8 days in the production vessel) for the Pareto optimal solutions for 5 shake flask scales}.
	\label{fig: spider_4obj_5sf_20runs}
\end{figure}

To obtain a visual overview for multiple objective criteria in one figure, a so-called spider plot (or net plot) can be used, which is shown in Figure~\ref{fig: spider_4obj_5sf_20runs}.
The horizontal axis shows the values of the deviation rate (on the right) and of the viability (on the left). The vertical axis shows the values of the seed train duration (above) and of titer (below). The aim of the optimization was to minimize seed train duration and deviation rate and to maximize viability and titer.
Each color (hyperplane) represents one of the Pareto optimal seed train configurations (based on the optimal combinations of filling volumes in shake flask scales). 
Since seed train duration and deviation rate should be minimal and titer and viability should be maximal, hyperplanes covering the lower left area would be desired. However, no such solution (hyperplane)  was obtained. The reason is that the optimization problem contains conflicting objective criteria, meaning that an improvement of one criterion leads to a degradation of another criterion. The here presented solutions are all non-dominated (compare to the green circles in the figures for 2 objective criteria).
For all shown solutions, deviation rate is rather low (4.9--7.3\%), the seed train duration lies between 521 and 562 hours and a titer of approximately 430--433 mg/L (assuming here a cell-specific production rate of $q_{\rm titer, max}=3.9\cdot10^{-10}$ mg\,cell$^{-1}$\, h$^{-1}$, as reported in \cite{Hernandez.2021}) and a viability of 52--53\% is reached after 8 days in the production vessel (here via batch-mode).
Of course, the obtained values depend a lot on the real process conditions (production bioreactor probably performed in fed-batch model) and the model parameter values obtained after model validation.  
However, the presented simulation example shall illustrate how the proposed approach can be applied for risk-based decision making under consideration of several criteria that should be optimal.

\subsection{Impact of performed iterations during Bayes optimization}\label{res: iterations}
\begin{figure}[h]
	\includegraphics[width=1\linewidth]{"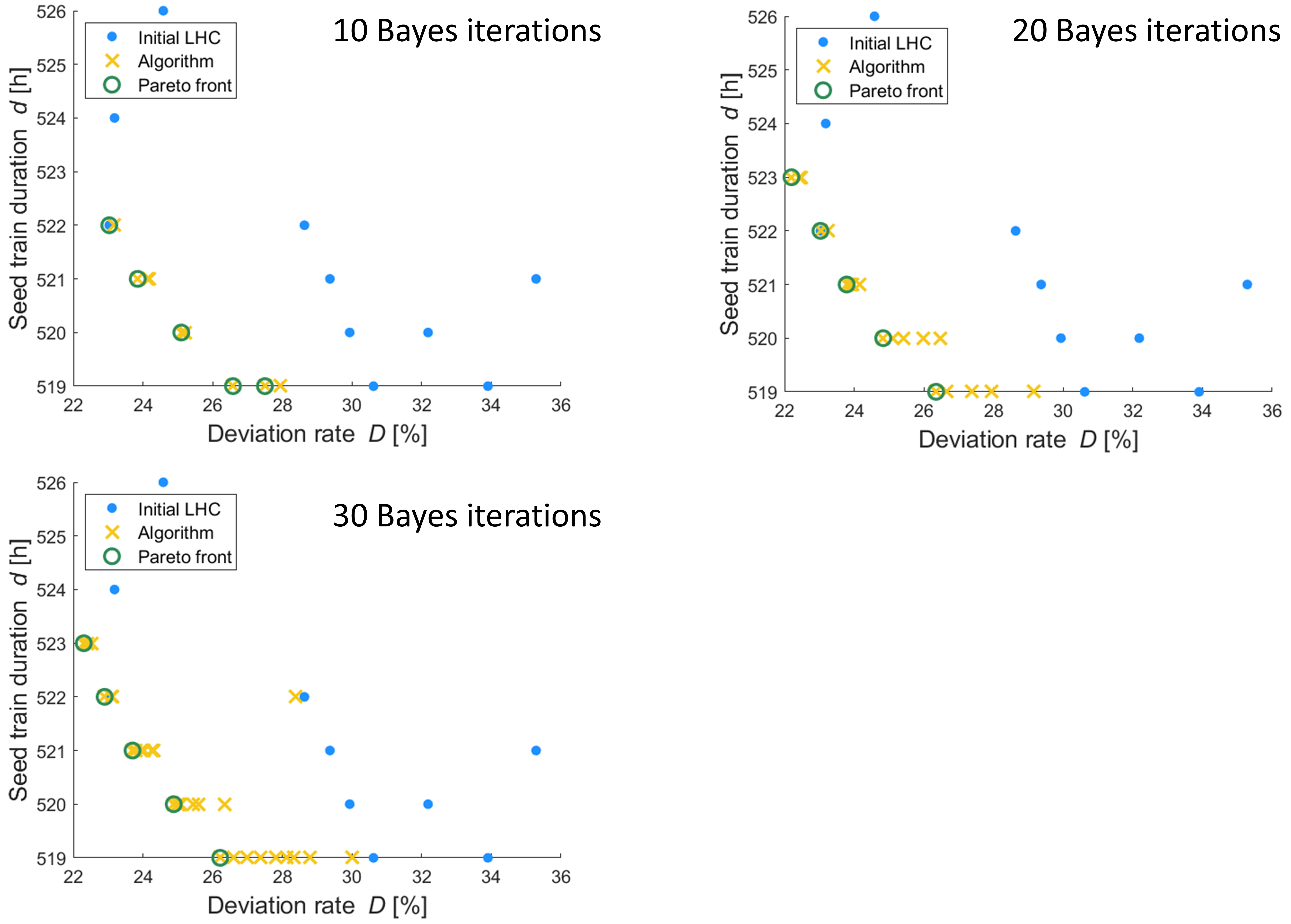"}
	\caption[]{Objective criterion {\em seed train duration} over objective criterion {\em deviation rate} obtained for evaluated points (optimizable variables, here combinations of 3 shake flask filling volumes) for 10 (top left), 20 (top right) and 30 (bottom left) Bayes iterations; blue dots: based on the initial Latin hypercube design (LHC); yellow crosses: based on the points proposed by the algorithm; green circles: Pareto optimal solutions (=\,Pareto front).}
\label{fig: comparison_10_20_30_bayes_iterations}
\end{figure}

For the example of 3 shake flask scales, (followed by 3 bioreactor scales) and optimizing filling volumes for all shake flask scales with respect to the two objective criteria: seed train duration and deviation rate, the number of performed iterations during the optimization procedure was varied. First, 10 initial points (combinations of filling volumes) distributed based on a Latin hypercube design were evaluated, followed by 10 Bayes iterations, which means that 10 times the algorithm updates the black box model (the Gaussian process), calculates the acquisition function and proposes the next point based on the outcome of this calculation. 
Then, the optimization was performed again for the same seed train setup but using 20 and then 30 Bayes iterations. The obtained solutions are illustrated in Figure~\ref{fig: comparison_10_20_30_bayes_iterations}.

Increasing the number of iterations from 10 to 20 helped to identify one solution that has not been discovered when running only 10 iterations. This can be seen when comparing the green circles in the diagram top left and the green circles in the diagram top right. The solution with $D\approx$\,22 and $d$\,=\,523 cannot been found in the diagram top left. \\
Increasing the number of iterations from 20 to 30 did not lead to an improved optimum as can be seen when comparing the green circles in Figure~\ref{fig: comparison_10_20_30_bayes_iterations} diagram top right and diagram bottom left. This underlines the efficiency of the Bayes optimization. In the present example only 10 initial points (distributed randomly according to a Latin hypercube design) and 20 Bayes optimization iteration steps were required to obtain the results which were confirmed when applying 30 iteration steps.

\section{Conclusion}
A concept has been developed to use process models in combination with algorithms for Bayes optimization using Gaussian processes to solve multi-objective optimization problems in the context of biopharmaceutical production processes.
To illustrate this approach, a relevant exemplary optimization problem was chosen and solved using the proposed method.

The goal was to find optimal combinations of filling volumes for the shake flask scales of a seed 577 train leading to a minimum deviation rate regarding viable cell densities and a minimum process 578 duration.
Compared to a non-optimized reference seed train, the optimized process showed much 579 lower deviation rates regarding viable cell densities (\textless~10\% instead of 41.7\%) using 5 or 4 shake flask 580 scales and seed train duration could be reduced by 56\,h from 576\,h to 520\,h.

Overall, it is shown that applying Bayes optimization to a multi-objective optimization function with several optimizable input variables and under a considerable amount of constraints, lead to revealing results with a low computational effort.
This approach provides the potential to be used in form of a decision tool, e.g.\ for the choice of an optimal and robust seed train design but also to further optimization tasks within process development. 

It should be noted that Bayes optimization and the corresponding computational modules could also be applied, even if no mechanistic process model is available, following a slightly different workflow. Instead of performing model-based in-silico experiments (process simulations), real lab experiments would be performed and fed back to update the black box model (here the Gaussian process). This adaptive procedure (also called Bayesian experimental design or experimental design with Bayesian optimization \cite{Greenhill.2020}) or further related optimization methods might be promising tools to support experimental planning, process characterization, process transfer or optimization of cell culture processes but they still require further research and being embedded in software solutions easy to use for operators.

\clearpage

\vspace{6pt} 

\section*{Acknowledgments}

{
	The authors would like to express special thanks to Christoph Posch (Novartis Technical Research \& Development) for the fruitful scientific exchange regarding the here presented case study. Moreover, we acknowledge support for the Open Access fees by Ostwestfalen-Lippe University of Applied Sciences and Arts (TH OWL) in the funding program Open Access Publishing.}

\section*{Abbreviations}
{The following abbreviations are used in this manuscript:\\

\noindent 
\begin{tabular}{@{}ll}
CHO 	& Chinese hamster ovary \\
EI		& Expected improvement \\
FDA		& Food and Drug Administration \\
GP  	& Gaussian process \\
LHS		& Latin hypercube sampling \\
ode		& Ordinary differential equations \\
SE		& Squared exponential \\
VCD		& Viable cell density \\
\end{tabular}}

\begin{table}
	\label{tab:list_of_symbols}    
	\begin{tabular}{ll}
		{\bf List of symbols} & \\
$\alpha$  & Risk aversion parameter (-) \\
$\mu$  & Cell-specific growth rate (h$^{-1}$) \\
$\mu_{\rm d}$  & Cell-specific death rate (h$^{-1}$ ) \\
$\mu_{\rm d,max}$  & Maximum cell-specific death rate (h$^{-1}$) \\
$\mu_{\rm d,min}$  & Minimum cell-specific death rate (h$^{-1}$) \\
$\mu_{\rm max}$	&   Maximum cell-specific growth rate (h$^{-1}$)\\
$\mu_{\rm ref}$	&   Reference maximum cell-specific growth rate (h$^{-1}$)\\
$\sigma^2$	& Variance \\
$c_{\rm Amm}$ ($c_{\rm Amm,0}$) & (Initial) ammonia concentration (mmol L$^{-1}$)\\
$c_{\rm Glc}$ ($c_{\rm Glc,0}$) & (Initial) glucose concentration (mmol L$^{-1}$)\\
$c_{\rm Gln}$ ($c_{\rm Gln,0}$) & (Initial) glutamine concentration (mmol L$^{-1}$) \\	
$c_{\rm Lac}$ ($c_{\rm Lac,0}$) & (Initial) lactate concentration (mmol L$^{-1}$)\\	
$c_{\rm titer}$ ($c_{\rm titer,0}$) & (Initial) volumetric titer (product concentration) (mg L$^{-1}$)\\	
$d$			& Dimension of the input space, number of optimizable variables \\
& =~seed train duration (h)\\
$D$			& Data, Deviation rate \\
${\rm E}(\cdot)$ 	& Expectation value \\
$f$			& Objective function (-) \\
$f_i$			& Component $i$ of a multidimensional objective function (-) \\	
$F_{\rm sample}$ & Change of volume due to sampling (L h$^{-1}$)\\
$i$   	& Running index (-) \\
$k$			& Covariance function \\
$K_{ {\rm Amm}}$  & Correction factor for ammonia uptake (-)\\
$K_{{\rm Lys}}$   &  Cell lysis constant (h$^{-1}$)\\
$K_{{\rm S,Glc}}$  & Monod kinetic constant for glucose (mmol L$^{-1}$) \\
$K_{{\rm S,Gln}}$ & Monod kinetic constant for glutamine (mmol L$^{-1}$) \\
$k_{{\rm Glc}}$ &   Monod kinetic constant for glucose uptake (mmol L$^{-1}$)\\
$k_{{\rm Gln}}$	&   Monod kinetic constant for glutamine uptake (mmol L$^{-1}$) \\
$m$	($m(\cdot)$)	& Mean (mean function)  \\
$n$				& Number of shake flasks (-) \\	
$N$				& Number of iterations (-) \\
$\mathcal{N}$				& Normal distribution (-) \\
$n_{\rm lhs}$   & Number of latin hypercube points (-) \\
$q_{Amm}$ ($q_{ {\rm Amm,uptake,max}}$) & (Maximum) cell-specific ammonia uptake rate (mmol cell$^{-1}$ h$^{-1}$) \\
$q_{ {\rm Glc}}$ ($q_{ {\rm Glc,max}}$) & (Maximum) cell-specific glucose uptake rate (mmol cell $^{-1}$ h$^{-1}$ )\\
$q_{ {\rm Gln}}$ ($q_{ {\rm Gln,max}}$) & (Maximum) cell-specific glutamine uptake rate (mmol cell $^{-1}$ h$^{-1}$ )\\
$q_{ {\rm Lac}}$ ($q_{ {\rm Lac,uptake,max}}$)  & (Maximum) cell-specific lactate uptake rate (mmol cell$^{-1}$ h$^{-1}$ )\\
$q_{ {\rm titer}}$ ($q_{\rm titer,max}$) &	(Maximum) cell-specific product production rate (mg cell$^{-1}$ h$^{-1}$)\\
$ \mathbb{R}$ 	& Set of real number \\
$t$ 	& Time (h)  \\
$T_{\rm p}$ 	& Set of feasible points in time for passaging \\
$U(\cdot) $		& Utility function \\
$V$	& Volume (L) \\
$V_i$	& Volume in shake flask scale $i$ \\
${\rm Var}(\cdot)$ 	& Variance \\
$x_c$ & Candidate point \\
$x$, $x'$	& Multidimensional points (vectors) of the input space \\
$x*$			& Argument that maximizes $f(s)$  \\
$X_{\rm t}$   & Total cell density (cells L$^{-1}$) \\	
$X_{\rm v}$   & Viable cell density (cells L$^{-1}$) \\
$X_{{\rm v},i}$   & Viable cell density at point in time with index $i$ (cells L$^{-1}$) \\
$\mathcal{X}$	& Input space \\
$y$ & Arbitrary function (-) \\
$Y$  & Arbitrary random variable (-) \\
$Y_{{\rm Amm/Gln}}$ & Kinetic production constant for ammonia (mmol mmol$^{-1}$)  \\
$Y_{{\rm Lac/Glc}}$ & Kinetic production constant for lactate (mmol mmol$^{-1}$) \\
	\end{tabular}
\end{table}

\clearpage
\appendix

\section{Supplementary material}
\unskip

\begin{table}[h]
	\footnotesize
	\caption[Mechanistic model for description of cell growth, cell death, substrate uptake, metabolite production and antibody production applicable to batch and fed-batch mode.]{Mechanistic model \cite{Frahm.2014, Kern.2016, HernandezRodriguez.2019,Moller.2019, Portner.2017} for description of cell growth, cell death, substrate uptake, metabolite production and antibody production applicable to batch and fed-batch mode.}
	\label{tab: model_eq_novartis}
	\centering
	\begin{tabular*}{\textwidth}{l|l}
		{\small Balance equations }& Kinetic equations \\
	\hline
		Biomass  &  \\
		$ \frac{dX_{\rm v}}{dt} = X_{\rm v} \cdot (\mu-\mu_{{\rm d}}) -\frac{F_{\rm Glc} + F_{\rm Gln}+F_{\rm Medium}}{V}\cdot X_{\rm v}$ & $ \mu = \mu_{\rm max} \cdot \frac{c_{{\rm Glc}} }{ c_{{\rm Glc}} + K_{{\rm S,Glc}}} \cdot \frac{ c_{{\rm Gln}} } {c_{{\rm Gln}} + K_{{\rm S,Gln}} } 
		$, if\, $t>t_{\rm Lag}$ \\
		\rule{0pt}{5ex}%
		& $ \mu = \mu_{\rm max} \cdot \frac{c_{{\rm Glc}} }{ c_{{\rm Glc}} + K_{{\rm S,Glc}}} \cdot \frac{ c_{{\rm Gln}} } {c_{{\rm Gln}} + K_{{\rm S,Gln}} } 
		- (1 - \frac{t}{t_{\rm Lag}}) \cdot a_{\rm Lag} \cdot \mu_{\rm max} $, \\
		& \qquad  if $t \leq t_{\rm Lag} $ \\
		$ \frac{dX_{\rm t} }{dt} = X_{\rm v} \cdot \mu - K_{{\rm Lys}} \cdot (X_{\rm t}-X_{\rm v})$ & $\mu_{{\rm d}}   = \mu_{{\rm d,min}} + \mu_{ {\rm d,max} } \cdot \frac{K_{{\rm S,Glc}} }{  K_{{\rm S,Glc}}  + c_{{\rm Glc}} } 
		\cdot \frac{K_{{\rm S,Gln}} }{  K_{{\rm S,Gln}}  + c_{{\rm Gln}} }   $ \\
		$ \quad \qquad -\frac{F_{\rm Glc} + F_{\rm Gln}+F_{\rm Medium}}{V} \cdot X_{\rm t}$  & \\
		\rule{0pt}{5ex}%
		Substrates &  \\ 
		$\frac{dc_{{\rm Glc}}}{dt} = -X_{\rm v} \cdot q_{{\rm Glc}} + \frac{F_{\rm Glc}}{V} \cdot c_{\rm Glc,F} +  \frac{F_{\rm Medium}}{V} \cdot c_{\rm Glc,Medium}$  & $q_{{\rm Glc}} = q_{ {\rm Glc,max}} \cdot \frac{ c_{{\rm Glc}} }{ c_{{\rm Glc}} + k_{{\rm Glc}}}  
		$ \\
		 \qquad \qquad $- \frac{F_{\rm Glc} + F_{\rm Gln}+F_{\rm Medium}}{V} \cdot c_{\rm Glc} $ & \\
		\rule{0pt}{3ex}%
		$\frac{dc_{{\rm Gln}}}{dt} = -X_{\rm v} \cdot q_{{\rm Gln}} + \frac{F_{\rm Gln}}{V} \cdot c_{\rm Gln,F} +  \frac{F_{\rm Medium}}{V} \cdot c_{\rm Gln,Medium}$
		& $ q_{{\rm Gln}} =  q_{ {\rm Gln,max}} \cdot \frac{  c_{{\rm Gln}} }{ c_{{\rm Gln}} + k_{{\rm Gln}}}$ \\
		\qquad \qquad $ -\frac{F_{\rm Glc} + F_{\rm Gln}+F_{\rm Medium}}{V} \cdot c_{\rm Gln}$ & \\
		\rule{0pt}{5ex}%
		Metabolites & \\
		$\frac{dc_{{\rm Lac}}}{dt} =  X_{\rm v} \cdot q_{ {\rm Lac}} -\frac{F_{\rm Glc} + F_{\rm Gln}+F_{\rm Medium}}{V} \cdot c_{\rm Lac}$ & $q_{ {\rm Lac}} =  Y_{{\rm Lac/Glc}} \cdot  q_{{\rm Glc}} \cdot \frac{c_{{\rm Glc}} }{ c_{{\rm Lac}} } - q_{ {\rm Lac,uptake}} \cdot \frac{ \mu_{\rm max} - \mu }{ \mu_{\rm max}} $ \\
		& \qquad \quad with \, $q_{\rm Lac,uptake} = 0$, if\, $c_{\rm Glc} > 0.5 {\small \rm \,mmol\,L}^{-1}$ \\
		& \qquad \quad with \, $q_{\rm Lac,uptake} = q_{\rm Lac,uptake,max}$, if
		$c_{\rm Glc} \leq 0.5 {\rm \,mmol\,L\,}^{-1}$ \\
		$\frac{dc_{{\rm Amm}}}{dt} =  X_{\rm v} \cdot q_{{\rm Amm}} -\frac{F_{\rm Glc} + F_{\rm Gln} + F_{\rm Medium}}{V} \cdot c_{\rm Amm}$ & $q_{{\rm Amm}} =  Y_{{\rm Amm/Gln}} \cdot q_{{\rm Gln}} \cdot \frac{ c_{{\rm Gln}}}{c_{{\rm Amm}}} $   \\
		& \qquad \quad $- K_{ {\rm Amm}} \cdot q_{ {\rm Amm,uptake,max}} \cdot     \frac{\mu_{\rm max} - \mu }{ \mu_{\rm max}} $  \\
		& \qquad \quad with $K_{\rm Amm} = 0$, if $ \quad (c_{\rm Gln} > c_{\rm Amm})$ \\
		& \qquad \quad with $ K_{\rm Amm} = 1\quad$, if $ \quad (c_{\rm Gln} \leq c_{\rm Amm}) \, \mbox{and} \, (\mu > \mu_d)$ \\
		& \qquad \quad with $K_{\rm Amm} = - k_{\rm Amm} \quad \mbox{(constant)}$, if $(\mu \leq \mu_{\rm d}) $ \\
		Product titer and volume & \\
		\rule{0pt}{3ex}%
		$\frac{dc_{{\rm titer}}}{dt} = X_{\rm v} \cdot q_{\rm titer} -\frac{F_{\rm Glc} + F_{\rm Gln}+F_{\rm Medium}}{V} \cdot c_{\rm titer}$ & $ q_{\rm titer} = q_{ {\rm titer,max}}$ \\
		\rule{0pt}{3ex}%
		$\frac{dV}{dt} = - F_{\rm Sample} + F_{\rm Glc} + F_{\rm Gln}+F_{\rm Medium}$ & \\
	\end{tabular*}
\end{table}

\clearpage

\section{Supplementary figures}

\subsection{Application to other cell lines with potentially higher and lower maximum growth rates}

\begin{figure}[h]
	\centering
	\begin{subfigure}[b]{0.3\textwidth}
		\includegraphics[width=\textwidth]{figures/Fig_pareto_front_5sf_mu100_2obj_lhs10_it20_appendix.png}
		\caption{5 sf, $\mu_{\rm,max,ref}$.}
		\caption*{ }
	\end{subfigure}%
	~ 
	\begin{subfigure}[b]{0.3\textwidth}
		\includegraphics[width=\textwidth]{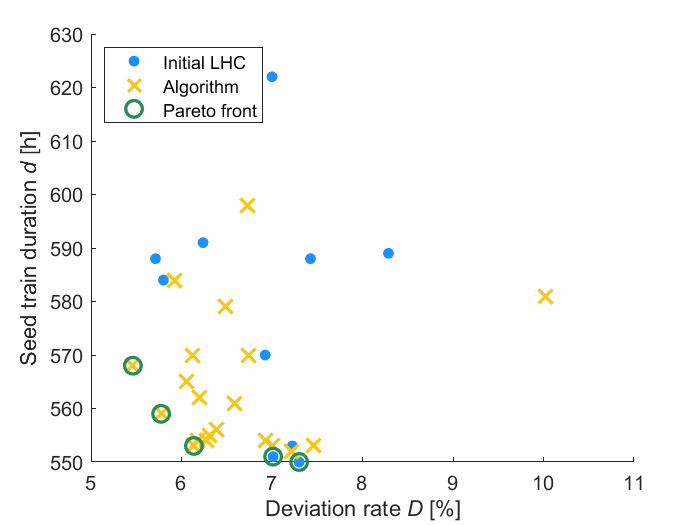}
		\caption{5 sf, $\mu_{\rm,max,95\%}$.}
		\caption*{ }
	\end{subfigure}
	\begin{subfigure}[b]{0.3\textwidth}
		\includegraphics[width=\textwidth]{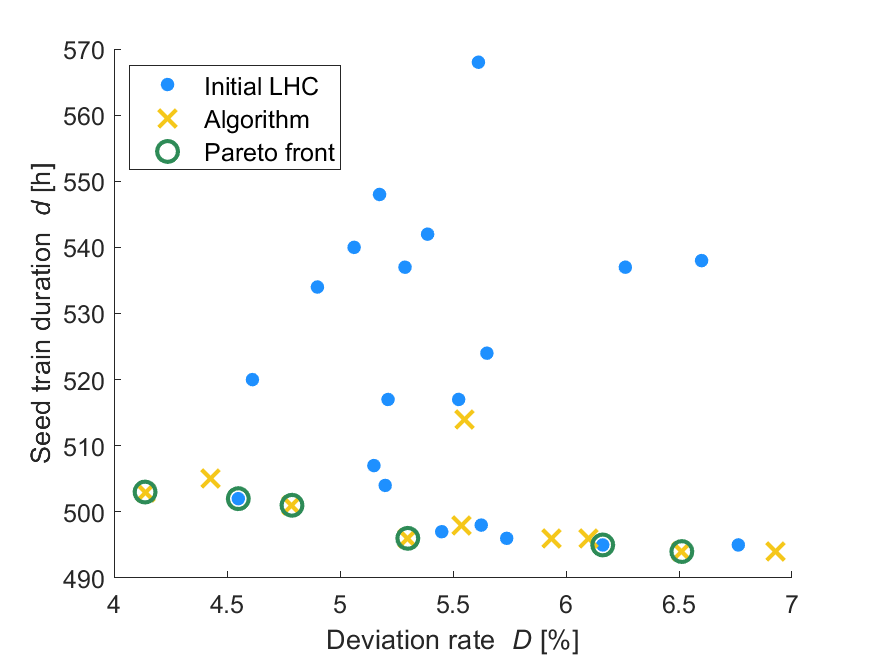}
		\caption{5 sf, $\mu_{\rm,max,105\%}$.}
		\caption*{ }
	\end{subfigure}
	\begin{subfigure}[b]{0.3\textwidth}
		\includegraphics[width=\textwidth]{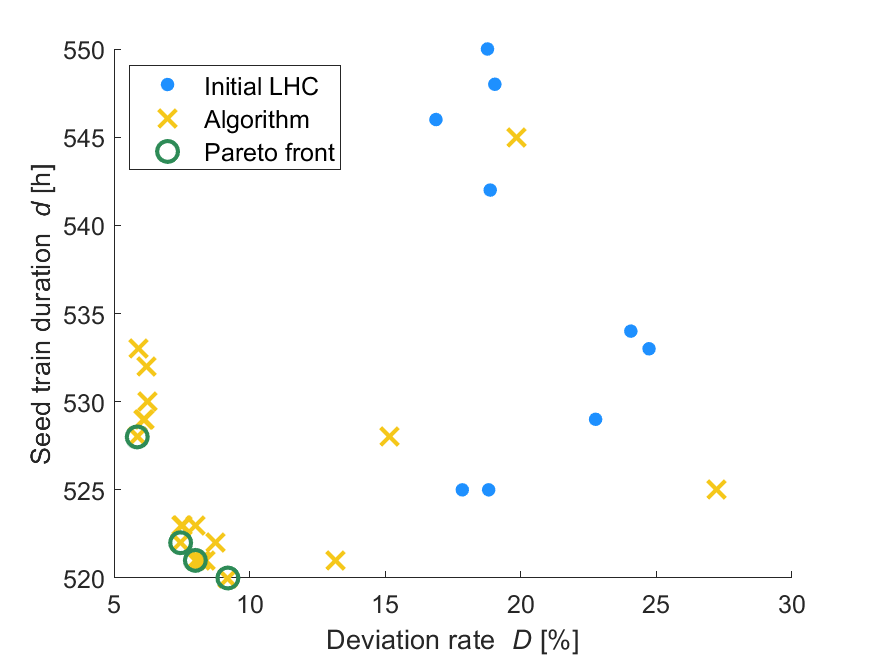}
		\caption{4 sf, $\mu_{\rm,max,ref}$.}
		\caption*{ }
	\end{subfigure}
	\begin{subfigure}[b]{0.3\textwidth}
		\includegraphics[width=\textwidth]{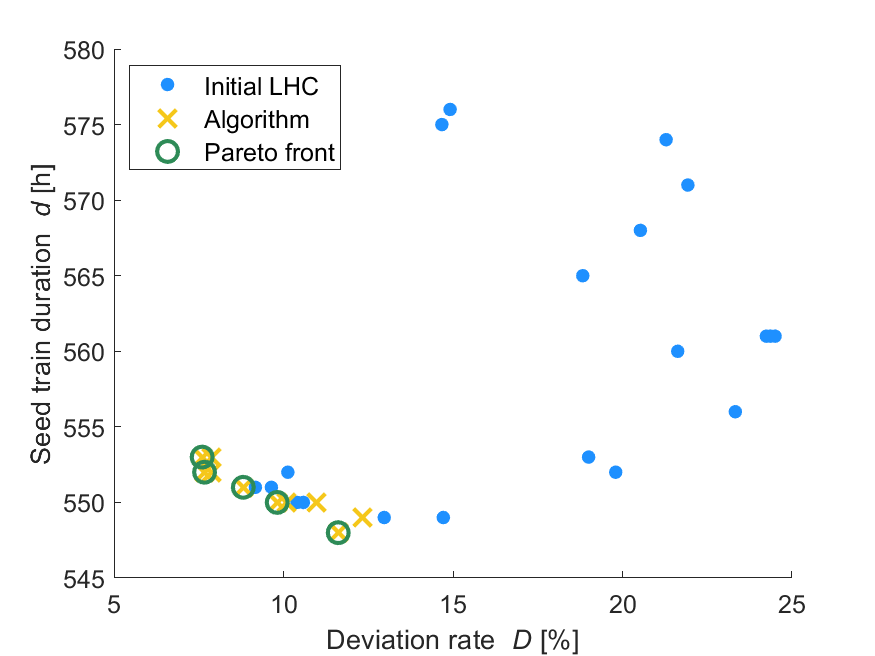}
		\caption{4 sf, $\mu_{\rm,max,95\%}$.}
		\caption*{ }
	\end{subfigure}
	\begin{subfigure}[b]{0.3\textwidth}
		\includegraphics[width=\textwidth]{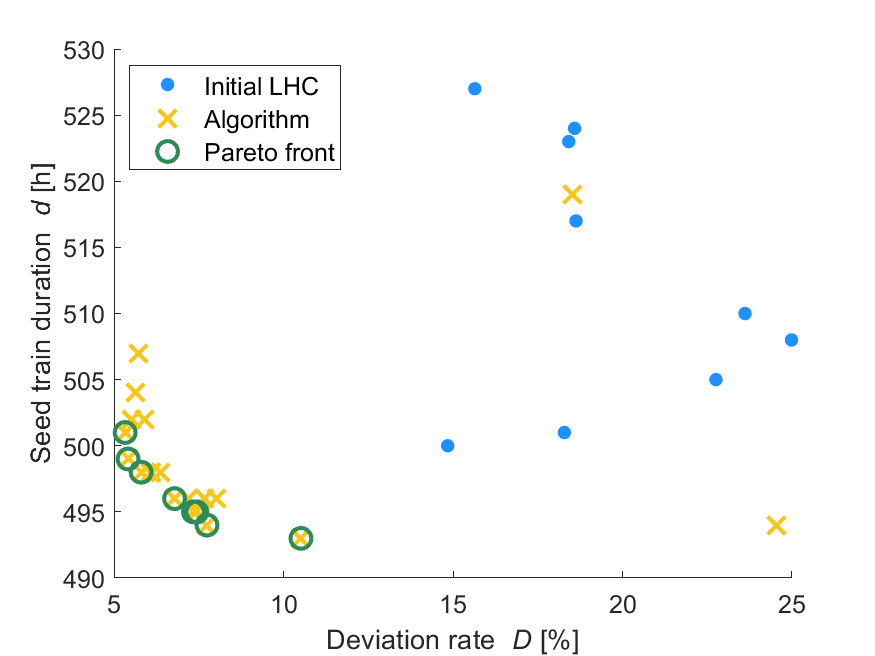}
		\caption{4 sf, $\mu_{\rm,max,105\%}$.}
		\caption*{ }
	\end{subfigure}
	\begin{subfigure}[b]{0.3\textwidth}
		\includegraphics[width=\textwidth]{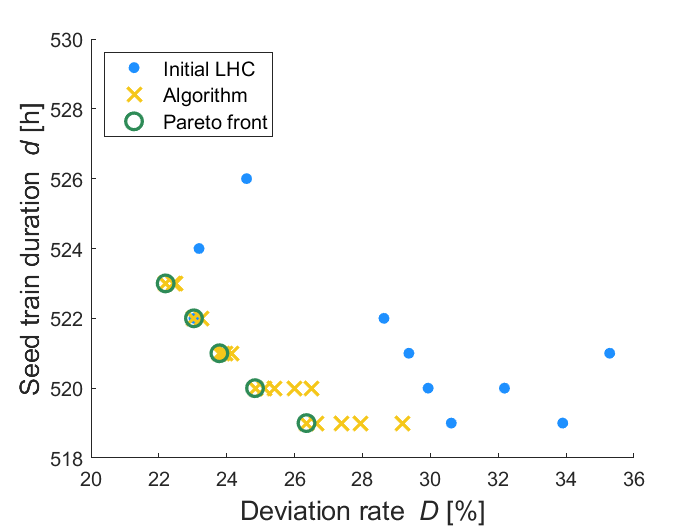}
		\caption{3 sf, $\mu_{\rm,max,ref}$.}
		\caption*{ }
	\end{subfigure}
	\begin{subfigure}[b]{0.3\textwidth}
		\includegraphics[width=\textwidth]{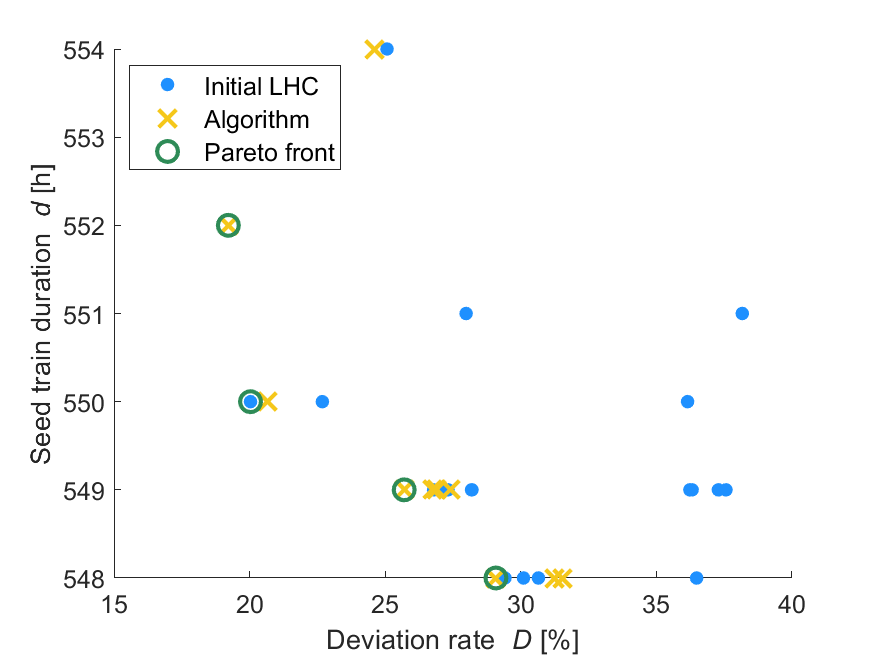}
		\caption{3 sf, $\mu_{\rm,max,95\%}$.}
		\caption*{ }
	\end{subfigure}
	\begin{subfigure}[b]{0.3\textwidth}
		\includegraphics[width=\textwidth]{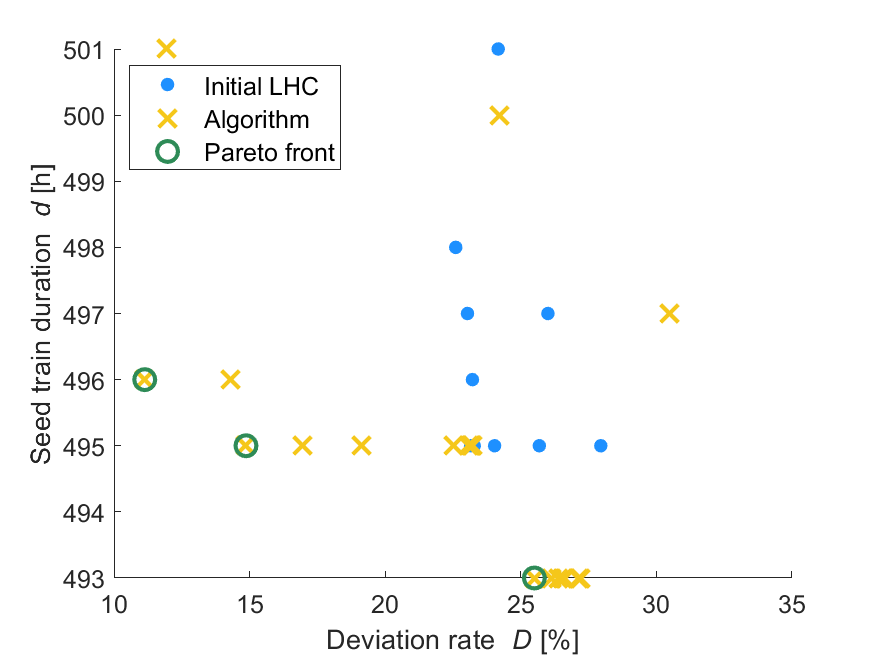}
		\caption{3 sf, $\mu_{\rm,max,105\%}$.}
		\caption*{ }
	\end{subfigure}
	\caption{Pareto solutions for 3, 4 and 5 shake flask scales and for three different growth rates, reference maximum growth rate (left column), a 5\% lower (middle column) and a 5\% higher growth rates (right column) showing the objective criterion seed train duration over objective criterion deviation rate, using 20 optimization iterations; Blue dots: based on the initial Latin hypercube (LHC) design; Yellow crosses: based on the proposed points (by the algorithm); Green circles: Pareto optimal solutions.}
	\label{fig: app_Pareto_solutions}
\end{figure}

\clearpage
\bibliographystyle{plain}

\end{document}